\documentclass[11pt]{article}
\usepackage{geometry}                
\usepackage{setspace}
\usepackage[parfill]{parskip}    
\usepackage{newtxtext}

\usepackage{amsfonts}
\usepackage{amsmath}
\usepackage{amsthm}
\newtheorem{theorem}{Theorem}[section]
\newtheorem{corollary}{Corollary}[section]
\newtheorem{lemma}{Lemma}[section]
\newtheorem{definition}{Definition}[section]
\newtheorem{assumption}{Assumption}[section]
\newtheorem{remark}{Remark}[section]

\usepackage[hidelinks]{hyperref} 
\usepackage{url}
\usepackage{booktabs} 
\usepackage{bm}
\usepackage{array}
\usepackage{multirow}
\usepackage{graphicx}
\usepackage{subcaption}

\usepackage{color}
\usepackage{float}
\usepackage{wrapfig}

\title{Robustness, Privacy, and Generalization of Adversarial Training}

\author{Fengxiang He~\thanks{These authors contributed equally.}~~\thanks{F. He, S. Fu, and D. Tao are with UBTECH Sydney AI Centre, School of Computer Science, Faculty of Engineering, the University of Sydney, Darlington NSW 2008, Australia. Email: \href{mailto:fengxiang.he@sydney.edu.au}{fengxiang.he@sydney.edu.au}, \href{mailto:shfu7008@uni.sydney.edu.au}{shfu7008@uni.sydney.edu.au}, and \href{mailto:dacheng.tao@sydney.edu.au}{dacheng.tao@sydney.edu.au}.} \and Shaopeng Fu~\footnotemark[1]~~\footnotemark[2] \and Bohan Wang~\footnotemark[1]~~\thanks{B. Wang was with School of Mathematical Sciences, University of Science and Technology of China, Hefei, Anhui, 230026, China, when this work was completed. He is currently with Microsoft Research Asia, Beijing, 100089, China. Email: \href{mailto:bhwangfy@gmail.com}{bhwangfy@gmail.com}.} \and Dacheng Tao~\footnotemark[2]}

\date{}

\begin{document}

\maketitle

\begin{abstract}
Adversarial training can considerably robustify deep neural networks to resist adversarial attacks. However, some works suggested that adversarial training might comprise the privacy-preserving and generalization abilities. This paper establishes and quantifies the privacy-robustness trade-off and generalization-robustness trade-off in adversarial training from both theoretical and empirical aspects. We first define a notion, {\it robustified intensity} to measure the robustness of an adversarial training algorithm. This measure can be approximate empirically by an asymptotically consistent empirical estimator, {\it empirical robustified intensity}. Based on the robustified intensity, we prove that (1) adversarial training is $(\varepsilon, \delta)$-differentially private, where the magnitude of the differential privacy has a positive correlation with the robustified intensity; and (2) the generalization error of adversarial training can be upper bounded by an $\mathcal O(\sqrt{\log N}/N)$ on-average bound and an $\mathcal O(1/\sqrt{N})$ high-probability bound, both of which have positive correlations with the robustified intensity. Additionally, our generalization bounds do not explicitly rely on the parameter size which would be prohibitively large in deep learning. Systematic experiments on standard datasets, CIFAR-10 and CIFAR-100, are in full agreement with our theories. The source code package is available at \url{https://github.com/fshp971/RPG}.
\end{abstract}

~~~~~~~~~~\textbf{Keywords:} adversarial training, adversarial robustness, privacy preservation, generalization.

\section{Introduction}
\label{sec:introduction}

Adversarial training \cite{dai2018adversarial, li2018second, baluja2018learning, zheng2019distributionally} can considerably improve the adversarial robustness of deep neural networks against adversarial examples \cite{biggio2013evasion, szegedy2013intriguing, goodfellow2014explaining, papernot2016limitations, wen2019geometry}. Specifically, adversarial training can be formulated as solving the following minimax-loss problem,
\begin{equation*}
    \min_{\theta} \frac{1}{N} \sum_{i=1}^N
    \max_{\Vert x_{i}^\prime - x_{i} \Vert \leq \rho}
   \ell(h_{\theta} (x_i^\prime), y_i),
\end{equation*}
where $h_{\theta}$ is the hypothesis parameterized by $\theta$, $N$ is the training sample size, $x_i$ is a feature, $y_i$ is the corresponding label, and $l$ is the loss function. Intuitively, adversarial training optimizes neural networks according to the performance on worst-case examples, which are most likely to be adversarial examples.

This paper studies how adversarial training would influence the privacy-preserving \cite{dwork2013s, dwork2014algorithmic} and generalization \cite{vapnik2013nature, mohri2018foundations} abilities, both of which are of profound importance in machine learning. Based on both theoretical and empirical evidence, we prove that:
\begin{quote}
{\it The minimax-based approach can hurt the privacy-preserving and generalization abilities, while it can enhance the adversarial robustness.}
\end{quote}


The first question raised is {\it how to measure adversarial robustness?} Two straightforward measures would be the accuracy on the adversarial examples and the radius $\rho$ in adversarial training. However, it might be difficult to develop theoretical foundations upon either of them.

In this paper, we define a new term, {\it robustified intensity}, to assess the adversarial robustness of a learning algorithm. It is defined as the difference in the gradient norm introduced by the adversarial training, measured by the division operation. We further define an empirical estimator, {\it empirical robustified intensity}, for practical utilization. We prove that empirical robustified intensity is asymptotically consistent with robustified intensity. A comprehensive empirical study demonstrates that there is a clear positive correlation between robustified intensity and adversarial accuracy. This implies that robustified intensity is an informative measure.

We then study the privacy-robustness relationship. Instead of optimizing the average performance on all training examples, adversarial training optimizes neural networks on worst-case examples. This forces the learned model more heavily relying on a small subset of the training sample set. Therefore, one may have a considerably increased chance to launch a successful {\it differential attack}, which first replaces one training example by a fake example, and then inference other training examples by the change of the output model. We prove that adversarial training is $(\varepsilon, \delta)$-differentially private when using stochastic gradient descent (SGD) to optimize the minimax loss. Further, the magnitude of both $\varepsilon$ and $\delta$ have a positive correlation with the robustified intensity, which is the first result that establishes the theoretical foundations for the privacy-robustness trade-off.

Based on the privacy preservation, we prove an $\mathcal O(\sqrt{\log N}/N)$ on-average generalization bound and an $\mathcal O(1/\sqrt{N})$ high-probability generalization bound for adversarial training, where $N$ is the training sample size. The two bounds are established based on a novel theorem linking algorithmic stability and differential privacy. Furthermore, our generalization bounds do not have any explicit dependence on the parameter size, which can be prohibitively large in deep learning. The only term that would rely on the model size, the norm of the gradient, is verified by the experiments to be small. 


From the empirical aspect, we conduct systematic experiments on two standard datasets, CIFAR-10 and CIFAR-100 \cite{krizhevsky2009learning}, with two different metric norms, $L_\infty$ and $L_2$, for adversarial training while strictly controlling irrelative variables. The privacy-preserving abilities are measured by the accuracies of the membership inference attack \cite{shokri2017membership, yeom2018privacy}. Meanwhile, the generalizabilities are measured by the difference between the training accuracies and the test accuracies. The membership inference attack accuracies, generalization gaps, and empirical robustified intensities of the models trained in various settings are collected for analysis. The empirical results are in full agreement with our hypotheses. The training code, learned models, and collected data are available at \url{https://github.com/fshp971/RPG}.

The rest of this paper is organized as follows. Section \ref{sec:review} reviews related works. Section \ref{sec:preliminaries} presents notations and preliminaries necessary to the following discussions. Section \ref{sec:robustness} defines the robustified intensity and its empirical estimator. Sections \ref{sec:privacy_robustness} and \ref{sec:generalization} establish the privacy-robustness relationship and generalization-robustness relationship, respectively. Section \ref{app:proofs} collects all the omitted proofs. Section \ref{sec:exp} presents implementation details of our experiments. Section \ref{sec:conclusion} concludes this paper.

\section{Background}
\label{sec:review}

This section reviews the background of this work.

\textbf{Deep learning theory.} Deep learning has been deployed successfully in many real-world scenarios. However, the theoretical foundations of deep learning are still elusive. For example, there is no explanation for how deep learning algorithms work, why they can succeed, when they would fail, and whether they would hurt society. Such deficiency in explainability questions the transparency and accountability of deep learning, and further undermines our confidence of deploying deep learning in security-critical application domains, such \cite{litjens2017survey, sun2006road}, medical diagnosis \cite{kulikowski1980artificial, silver2016mastering}, and drug discovery \cite{chen2018rise}. Many works have emerged to establish the theoretical foundations of deep learning via VC dimension \cite{harvey2017nearly}, Rademacher complexity \cite{golowich2017size, bartlett2017spectrally}, covering number \cite{bartlett2017spectrally}, Fisher-Rao norm \cite{liang2019fisher, tu2020understanding}, PAC-Bayesian framework \cite{neyshabur2017pac}, algorithmic stability \cite{hardt2016train, kuzborskij2018data, verma2019stability}, and the dynamics of stochastic gradient descent or its variants \cite{mandt2017stochastic, mou2018generalization, he2019control}. Please see more related works in surveys \cite{e2020towards, he2020recent, poggio2020theoretical}. This work is committed to establishing theoretical foundations of privacy, generalization, adversarial attack in deep learning, all of which have profound importance in enhancing the explainability, transparency, and accountability of deep models.

\textbf{Generalization.} Good generalization guarantees that an algorithm learns the underlying patterns in training data rather than just memorize the data. In this way, good generalization abilities provide confidence that the models trained on existing data can be applied to similar but unseen scenarios. Three major approaches in analyzing the generalizability are seen in the literature: (1) generalization bounds based on the hypothesis complexity, including VC dimension \cite{blumer1989learnability, vapnik2006estimation}, Rademacher complexity \cite{koltchinskii2000rademacher, koltchinskii2001rademacher, bartlett2002rademacher}, and covering number \cite{dudley1967sizes, haussler1995sphere}. The results are usually obtained via concentration inequalities. They also suggest controlling the model size to secure the generalizability, which is no longer valid in deep learning; (2) generalization bounds based on the algorithmic stability \cite{rogers1978finite, bousquet2002stability, xu2011sparse}. The results in this stream follow the motivation that learning algorithms robust to small disturbances in input data usually have good generalizability; and (3) generalization bounds in the PAC-Bayes framework \cite{mcallester1999pac, mcallester1999some}. The results are obtained based on information-theoretical versions of concentration inequalities.

\textbf{Privacy preservation.} Deep learning has become a dominant player in many real-world application areas, including financial services \cite{fischer2018deep}, healthcare \cite{wang2012framework}, and biometric authentication \cite{snelick2005large}, where massive personal data has been collected. However, an increasing number of privacy breaches have been reported. An infamous scandal in 2018 shocked people that Cambridge Analytics harvested large amounts of personal data without consent for political advertising. The customers are fed meticulously selected advertisement to promote specific politicians and agendas. It sheds lights to the prohibitive reality that machine learning algorithms can quietly navigate consumers' choice by the data that was supposed to be private.

A popular measure for the privacy-preserving ability is differential privacy \cite{dwork2014algorithmic} based on the privacy loss; please see for more details. The magnitude of differential privacy $(\varepsilon, \delta)$ represents the ability to resist  {\it differential attacks} that employing fake data to steal private information in the training data. Many variants of differential privacy have emerged to date: (1) concentration differential privacy is proposed under assumptions that the distribution of the privacy loss is sub-Gaussian \cite{dwork2016concentrated, bun2016concentrated}; (2) mutual-information differential privacy replace the division operation in difference privacy by mutual information \cite{cuff2016differential, wang2016relation, liao2017hypothesis}; (3) KL differential privacy replace the mutual information by the KL divergence \cite{chaudhuri2019capacity}; (4) similarly, R\'enyi differential privacy replaces the KL divergence by R\'enyi divergence further \cite{mironov2017renyi, geumlek2017renyi}. Abadi {et al.} \cite{abadi2016deep} and Arachchige {et al.} \cite{arachchige2019local} have also studied the privacy preservation of deep learning. 

\textbf{Adversarial robustness.} Many works suggest that adversarial examples are widespread in the feature spaces of deep models \cite{biggio2013evasion, szegedy2013intriguing, goodfellow2014explaining, papernot2016limitations, chen2020universal, karim2020adversarial}. Specifically, for (almost) any training example, one can find an adversarial example closed to it but the neural network assigns the adversarial example to a different class. Thus, one can slightly modify an example to fool a neural network. This would expose deep learning-based systems to adversarial attacks \cite{nguyen2015deep, kurakin2016adversarial, madry2018towards, papernot2017practical, gilmer2018motivating}. Many defence strategies \cite{dai2018adversarial, li2018second, baluja2018learning, zheng2019distributionally, wang2020hamiltonian} can increase the robustness of neural networks to adversarial attacks \cite{nguyen2015deep, kurakin2016adversarial, madry2018towards, papernot2017practical, gilmer2018motivating}. However, it is also reported that these approaches would undermine privacy preservation and generalization.

\textbf{Generalization-robustness relationship.} Some works have studied the trade-off between generalization and robustness. Tsipras {\it et al.} \cite{tsipras2019robustness} prove the existence of a trade-off between the standard accuracy of a model and its robustness to adversarial perturbations.  Sun {\it et al.} \cite{sun2019towards} prove that adversarial training needs more training data to achieve the same test accuracy as standard ERM. Nakkiran \cite{nakkiran2019adversarial} suggest that ``robust classification may require more complex classifiers ({\it i.e.}, more capacity) than standard classification''. Additionally, they prove a quantitative trade-off between the robustness and standard accuracy for simple classifiers. Three $\mathcal O(1 / \sqrt{N})$  generalization bounds are given by \cite{yin2019rademacher, khim2018adversarial, tu2019theoretical}, which are based on the Rademacher complexity and covering number of the hypothesis space. A detailed comparison of the tightness is given in Section \ref{sec:generalization}. Schmidt {\it et al.} \cite{schmidt2018adversarially} prove that the hypothesis complexity of models learned by adversarial training is larger than those learned by empirical risk minimization (ERM), which is also verified empirically. However, the existing results relying on hypothesis capacity/complexity of neural networks, which are prohibitively large. Our paper proposes two novel generalization bounds at rate $\mathcal O(\sqrt{\log N}/N)$ and $\mathcal O(1/\sqrt{N})$, respectively, without explicitly relying on the capacity/complexity. Instead, gradient norm, the only factor in our bounds that could depend on the parameter size, is verified to be considerably small by experiments.

\textbf{Robustness-privacy relationship.} There have been initial attempts to study the robustness-privacy relationship. Some works suggest that differentially private machine learning algorithms are robust to adversarial examples \cite{lecuyer2018connection, lecuyer2019certified}. Pinot {\it et al.} \cite{pinot2019unified} define two terms, adversarial robustness and generalized adversarial robustness, to express the robustness to adversarial examples, which are similar to the differential privacy and its variants. The paper then argues that the two new terms are equivalent to R\'enyi differential privacy, but without theoretical proof. Phan {\it et al.} \cite{phan2019preserving} design algorithms with both theoretical guarantees in the privacy-preserving ability and adversarial robustness. Song {\it et al.} \cite{song2019privacy} conduct comprehensive experiments to investigate the relationship between robustness and privacy, with the results suggesting that adversarial training has privacy risks. However, there is so far no theoretical foundation has been established to discover the robustness-privacy relationship.

\section{Notations}
\label{sec:preliminaries}

Suppose $S = \{(x_1, y_1), \ldots, (x_N, y_N) | x_i \in \mathbb R^{d_X}, y_i \in \mathbb R^{d_Y}, i = 1, \ldots, N\}$ is a sample set, where $d_X$ and $d_Y$ are the dimension of the feature $X$ and the label $Y$, respectively. For the brevity, we define $z_i = (x_i, y_i)$, which is an independent and identically distributed (i.i.d.) observation of variable $Z = (X, Y) \in \mathcal Z$.

Differential privacy measures the ability of preserving privacy \cite{dwork2014algorithmic}. A stochastic algorithm $\mathcal{A}$ is called ($\varepsilon,\delta$)-differentially private, if for any subset $B \subset \mathcal H$ and any neighboring sample set pair $S$ and $S'$ which are different by only one example, we have
		\begin{equation}
		\label{eq:dp}
		\log \left[ \frac{\mathbb P_{\mathcal{A}(S)}(\mathcal{A}(S)\in B) - \delta}{\mathbb P_{\mathcal{A}(S')}(\mathcal{A}(S')\in B)} \right] \le \varepsilon.
		\end{equation}
Algorithm $\mathcal{A}$ is also called $\varepsilon$-differentially private, if it is $(\varepsilon, 0)$-differentially private. Differential privacy controls the privacy loss (cf. \cite{dwork2014algorithmic}, p.18) defined as below,
\begin{equation*}
\log \left[ \frac{\mathbb P_{\mathcal{A}(S)}(\mathcal{A}(S)\in B)}{\mathbb P_{\mathcal{A}(S')}(\mathcal{A}(S')\in B)} \right].
\end{equation*}
		
For the hypothesis $\mathcal A(S)$ learned by an algorithm $\mathcal A$ on the training sample set $S$, the expected risk $\mathcal R(\mathcal A(S))$ and empirical risk $\hat{\mathcal R}(\mathcal A(S))$ of the algorithm $\mathcal A$ are defined as follows,
\begin{gather*}
	 \mathcal R(\mathcal A(S)) = \mathbb E_{Z} \ell(\mathcal{A}(S), Z),\\
	\hat{\mathcal R}_S(\mathcal A(S)) = \frac{1}{N} \sum_{i=1}^N \ell(\mathbf{\mathcal A}(S), z_i).
\end{gather*}
It worths noting that the randomness of $\mathcal A(S)$ can come from both the stochastic algorithm $\mathcal A$ and the training sample set $S$. Then, the generalization error is defined as the difference between the expected risk and empirical risk, whose upper bound is called the generalization bound.

Learning algorithms usually solve the following empirical risk minimization (ERM) problem to approach the optimal hypothesis,
\begin{equation*}
\min_{\theta} \hat{\mathcal R}_S(\theta) = \min_{\theta} \frac{1}{N} \sum_{i=1}^N\ell(h_{\theta} (x_i), y_i).
\end{equation*}
We usually employ stochastic gradient-based optimizers for ERM in deep learning. Popular options of stochastic gradient-based optimizers include stochastic gradient descent (SGD) \cite{robbins1951stochastic}, momentum \cite{nesterov1983method, tseng1998incremental}, and Adam \cite{kingma2014adam}. For the brevity, we analyze SGD in this paper. The analysis for other stochastic gradient-based optimizers is similar.

Suppose $\mathcal B$ is a mini batch randomly drawn from the training sample set $S$. Then, the stochastic gradient on $\mathcal B$ is as follows,
\begin{equation*}
\hat g^\mathrm{ERM} (\theta) = \frac{1}{|\mathcal{B}|} \sum_{(x_i, y_i) \in \mathcal B} \nabla_{\theta}\ell(h_{\theta} (x_i), y_i).
\end{equation*}
In the $t$-th iteration, the weight is updated as follows,
\begin{equation*}
\theta^\mathrm{ERM}_{t+1} = \theta^\mathrm{ERM}_t - \eta_t \hat g^\mathrm{ERM} (\theta^\mathrm{ERM}_t),
\end{equation*}
where $\theta^\mathrm{ERM}_t$ is the weight vector in the $t$-th iteration and $\eta_t$ is the corresponding learning rate.

Meanwhile, adversarial training employs SGD to solve the following minimax problem,
\begin{equation}
\label{eq:adversarial_risk}
\min_{\theta} \hat{\mathcal R}^A_S(\theta) = \min_{\theta} \frac{1}{N} \sum_{i=1}^N \max_{\Vert x_{i}^\prime - x_{i} \Vert \leq \rho}\ell(h_{\theta} (x_i^\prime), y_i),
\end{equation}
where $\rho$ is the radius of the ball centered at the example $(x_i, y_i)$. Here, we call $\hat{\mathcal R}^A_S(\theta)$ adversarial empirical risk. Correspondingly, the stochastic gradient on a mini batch $\mathcal B$ and the weight update are calculated as below,
\begin{gather}
\label{eq:adversarial_gradient}
\hat g^A (\theta) = \frac{1}{|\mathcal B|} \sum_{(x_i, y_i) \in \mathcal B} \nabla_{\theta} \max_{\Vert x_{i}^\prime - x_{i} \Vert \leq \rho}\ell(h_{\theta} (x_i^\prime), y_i), \nonumber\\
\theta^A_{t + 1} = \theta^A_t - \eta_t \hat g^A (\theta^A_t).
\end{gather}

\section{Measurement of robustness}
\label{sec:robustness}

This section presents a new term, {\it robustified intensity}, to measure the adversarial robustness. We also design an asymptotically consistent empirical estimator, {\it empirical robustified intensity}, to empirically approximate robustified intensity.

\subsection{Robustified intensity}

We first define {\it single-iteration robustified intensity} as follows,

\begin{definition}[Single-iteration robustified intensity]
\label{def:ri_single}
The single-iteration robustified intensity of the $t$-th iteration in adversarial training (eq. \ref{eq:adversarial_risk}) is defined to be

\begin{small}
\begin{equation}
\label{eq:ri}
I_t = \frac{ \max_{(x, y) \in \mathcal Z} \left \| \left. \nabla_{\theta} \max_{\Vert x^\prime - x \Vert \leq \rho}\ell(h_{\theta} (x^\prime), y) \right|_{\theta=\theta^A_t} \right \|}{\max_{(x, y) \in \mathcal Z} \left\|  \left.  \nabla_{\theta}\ell(h_{\theta} (x), y) \right|_{\theta=\theta^\mathrm{ERM}_t} \right \|},
\end{equation}
\end{small}
where $x'$ is arbitrary in the input space subject to the condition $\|x' -x\| \le \rho$, $\| \cdot \|$ is the norm on the parameter space, and $\theta^A_t$ and $\theta^\mathrm{ERM}_t$ are the parameters in the $t$-th iteration of adversarial training and ERM, respectively.
\end{definition}

For brevity, we term the nominator and the denominator as $L^A_t$ and $L^\mathrm{ERM}_t$, respectively. Thus, $I_t = \frac{L^A_t}{L^\mathrm{ERM}_t}$. In adversarial training, $L^A_t$ is usually larger than $L^\mathrm{ERM}_t$ and thus $I_t > 1$.

We then extend our measure for adversarial robustness to whole training procedures. Suppose there are $T$ iterations in an adversarial training process, and the corresponding single-iteration robustified intensities are $I_1, \cdots, I_T$, respectively. 
We define a {\it robustified intensity} for the whole algorithm based on the single-iteration robustified intensities as below. 

\begin{definition}[Robustified intensity]
\label{def:ri}
Suppose $T$ iterations exist in an adversarial training procedure. Then, the robustified intensity for this procedure is defined to be
\begin{gather*}
	I_{1:T} = \left(\frac{1}{T} \sum_{t=1}^T I_t^4\right)^{\frac{1}{4}},
\end{gather*}
where $I_t$ is the single-iteration robustified intensity of the $t$-th interation.
\end{definition}

\begin{remark}
The fourth-order exponential operations in Definition \ref{def:ri} are designed for the convenience in establishing the privacy-robustness relationship. Moreover, our experiments show that such designing may not compromise the efficiency in evaluating the adversarial robustness.
\end{remark}

\subsection{How to empirically estimate robustified intensity?}

Calculating the robustified intensity is technically impossible as it involves searching the maximal value of gradient norms across the Euclidean space. We thus define {\it empirical single-iteration robustified intensity} and {\it empirical robustified intensity} to empirically estimate their theoretical counterparts.

\begin{definition}[Empirical single-iteration robustified intensity]
\label{def:empirical_ri_single}
The empirical single-iteration robustified intensity of the $t$-th iteration in adversarial training (eq. \ref{eq:adversarial_risk}) is defined to be
\begin{small}
\begin{equation*}
\hat I_t = \frac{ \left. \max_{(x_i,y_i)\in\mathcal B} \left \| \nabla_{\theta} \max_{\Vert x_{i}^\prime - x_{i} \Vert \leq \rho}\ell(h_{\theta} (x_i^\prime), y_i) \right|_{\theta=\theta^A_t} \right \|}{\max_{(x_i,y_i)\in\mathcal B} \left \| \left. \nabla_{\theta}\ell(h_{\theta} (x_i), y_i) \right|_{\theta=\theta^\mathrm{ERM}_t} \right \|},
\end{equation*}
\end{small}
where $\mathcal{B}$ is a mini batch sub-sampled from the training sample set $S$ and $\| \cdot \|$ is a norm defined in the space of the gradient.
\end{definition}

\begin{definition}[Empirical robustified intensity]
\label{def:empirical_ri}
Suppose $T$ iterations exist in an adversarial training procedure. Then, the empirical robustified intensity for this procedure is defined to be
\begin{gather*}
	\hat I_{1:T} = \left(\frac{1}{T} \sum_{t=1}^T \hat I_t^4\right)^{\frac{1}{4}},
\end{gather*}
where $\hat I_t$ is the empirical single-iteration robustified intensity of the $t$-th interation.
\end{definition}

\begin{figure*}[t]
    \begin{subfigure}{0.49\linewidth}
        \includegraphics[width=0.49\linewidth]{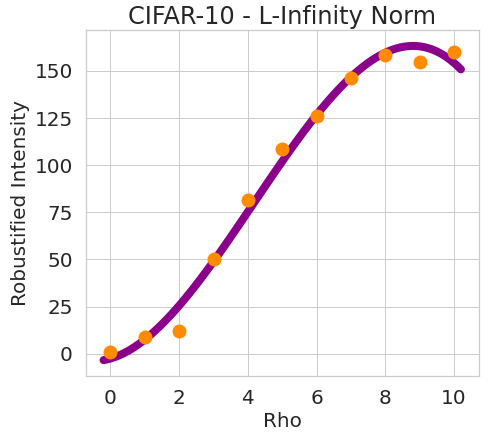}
        \includegraphics[width=0.49\linewidth]{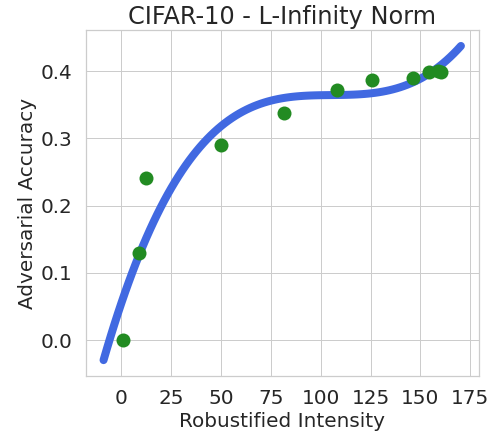}
        \subcaption{CIFAR-10, $L_\infty$ norm.}
    \end{subfigure}
    \begin{subfigure}{0.49\linewidth}
        \includegraphics[width=0.49\linewidth]{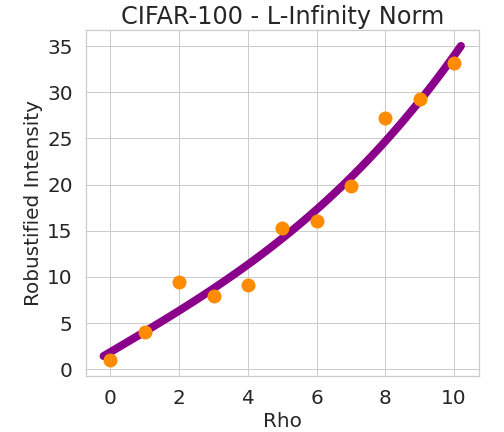}
        \includegraphics[width=0.49\linewidth]{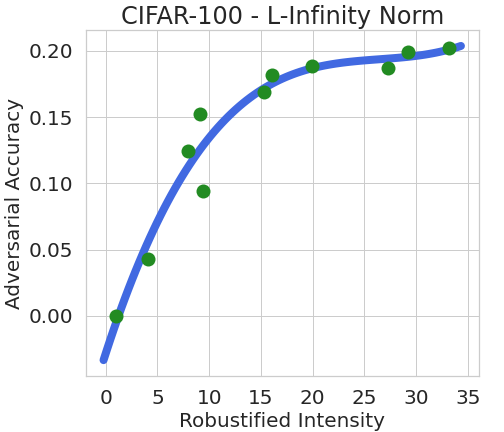}
        \subcaption{CIFAR-100, $L_\infty$ norm.}
	\end{subfigure}

    \begin{subfigure}{0.49\linewidth}
        \includegraphics[width=0.49\linewidth]{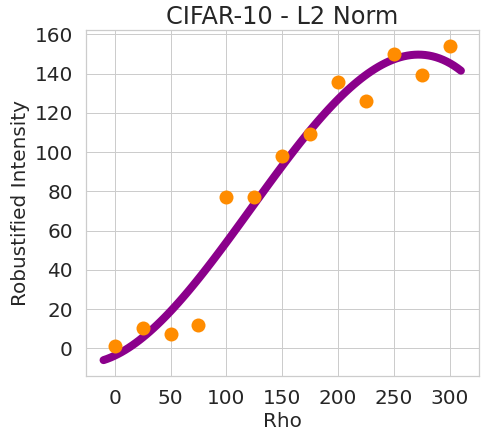}
        \includegraphics[width=0.49\linewidth]{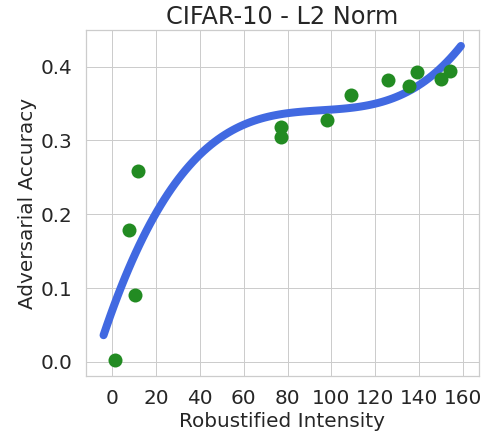}
        \subcaption{CIFAR-10, $L_2$ norm.}
    \end{subfigure}
    \begin{subfigure}{0.49\linewidth}
        \includegraphics[width=0.49\linewidth]{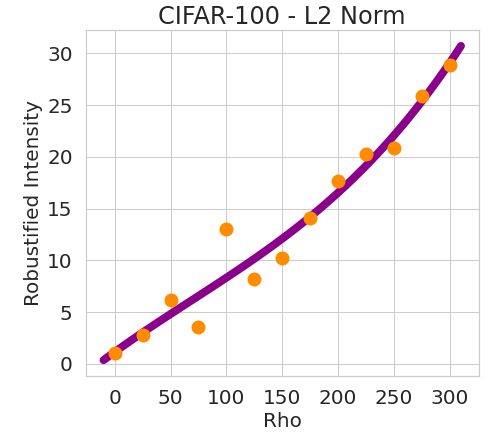}
        \includegraphics[width=0.49\linewidth]{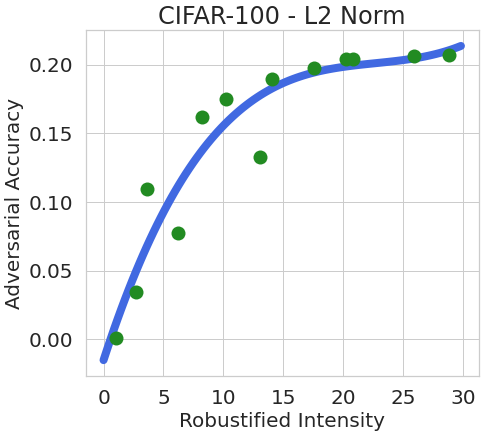}
        \subcaption{CIFAR-100, $L_2$ norm.}
    \end{subfigure}
    \caption{We conduct experiments on CIFAR-10 and CIFAR-100 with two different adversarial training metric norms, $L_\infty$ norm and $L_2$ norm. For each experiment, we draw two plots: (1) robustified intensity vs. radius $\rho$; and (2) adversarial accuracy vs. robustified intensity. Fourth-order polynomial regression is used for curve fitting in each figure. robustified intensity has positive correlations with both radius $\rho$ and adversarial accuracy, which demonstrates that robustified intensity is an informative robustness measurement.}
    \label{fig:robustness_intensity}
\end{figure*}

We can prove that both empirical estimators asymptotically consistent with their theoretical counterparts. The proofs need one mild assumption as below.

\begin{assumption}
		\label{assum:continuity}
		The gradient of loss function $\nabla_{\theta}\ell(\theta,z) \in C^0(\mathcal{Z})$; {\it i.e.}, for any hypothesis $h_\theta\in \mathcal{H}$, $\nabla
		_\theta \ell(h_\theta,z)$ is continuous with respect to example $z$.
\end{assumption}
	
The continuity with respect to the data $z$ can be easily meet in practice. Then, the asymptotical consistency is presented in the following two theorems.

\begin{theorem}[Asymptotic consistency of empirical single-iteration robustified intensity]
	\label{thm:asymtotic}
		Suppose the empirical single-iteration robustified intensity in the $t$-th training iteration is ${\hat I}^{(\tau)}_t$ where the batch size is $\tau$. Then, the ${\hat I}^{(\tau)}_t$ is an unbiased estimator for the  single-iteration robustified intensity $I_t$; {\it i.e.}, $\lim_{\tau \to \infty} {\hat I}^{(\tau)}_t = I_t$.
\end{theorem}

\begin{theorem}[Asymptotic consistency of empirical robustified intensity]
	\label{thm:asymtotic}
		Suppose the empirical robustified intensity in the $t$-th training iteration is ${\hat I}^{(\tau)}_{1:T}$ where the batch size is $\tau$. Then, the ${\hat I}^{(\tau)}_{1:T}$ is an unbiased estimator for the robustified intensity $I_t$; {\it i.e.}, $\lim_{\tau \to \infty} {\hat I}^{(\tau)}_{1:T} = I_{1:T}$.
	\end{theorem}
	
These two theorems secure that when the batch size for estimation is sufficiently large, the empirical single-iteration robustified intensity rigorously equals to the single-iteration robustified intensity. The proofs are novel and technically non-trivial. Please see details in Section \ref{app:asymtotic}.

\begin{figure*}[t]
    \begin{subfigure}[t]{0.23\linewidth}
        \includegraphics[width=\linewidth]{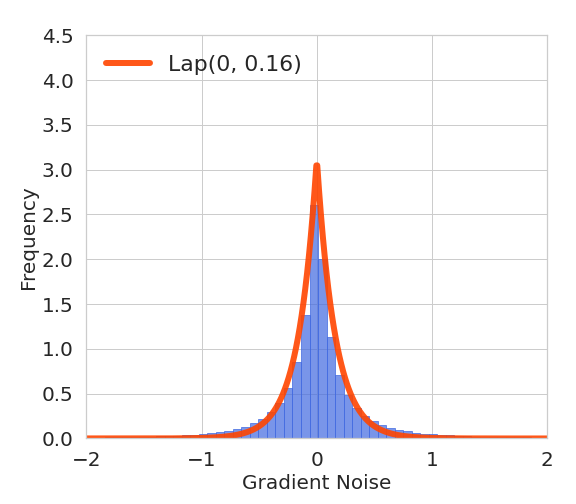}
        \subcaption{CIFAR-10, $40,000$ iterations, ERM, Lap$(0,0.16)$.}
    \end{subfigure}
    \hspace{2mm}
    \begin{subfigure}[t]{0.23\linewidth}
        \includegraphics[width=\linewidth]{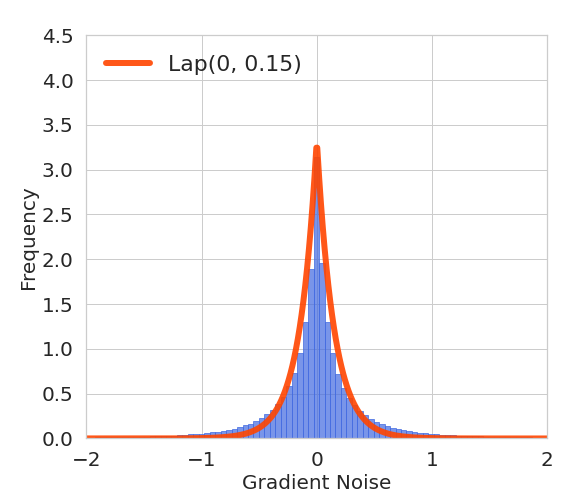}
        \subcaption{CIFAR-10, $80,000$ iterations, ERM, Lap$(0,0.15)$.}
    \end{subfigure}
    \hspace{2mm}
    \begin{subfigure}[t]{0.23\linewidth}
        \includegraphics[width=\linewidth]{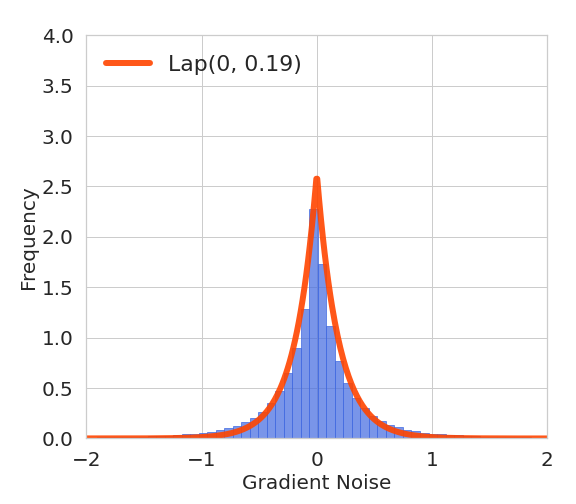}
        \subcaption{CIFAR-100, $40,000$ iterations, ERM, Lap$(0,0.19)$.}
    \end{subfigure}
    \hspace{2mm}
    \begin{subfigure}[t]{0.23\linewidth}
        \includegraphics[width=\linewidth]{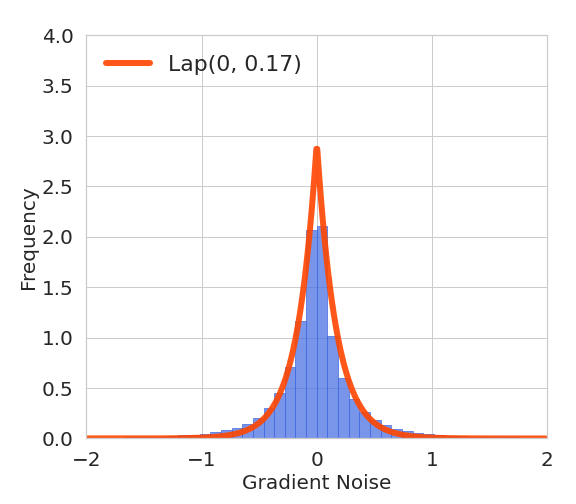}
        \subcaption{CIFAR-100, $80,000$ iterations, ERM, Lap$(0,0.17)$.}
    \end{subfigure}

    \begin{subfigure}[t]{0.23\linewidth}
        \includegraphics[width=\linewidth]{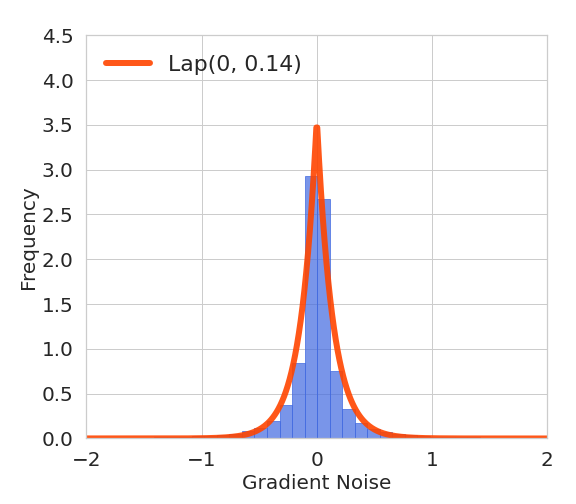}
        \subcaption{CIFAR-10, $80,000$ iterations, $L_\infty$ norm, $\rho=\frac{5}{255}$, Lap$(0,0.14)$.}
    \end{subfigure}
    \hspace{2mm}
    \begin{subfigure}[t]{0.23\linewidth}
        \includegraphics[width=\linewidth]{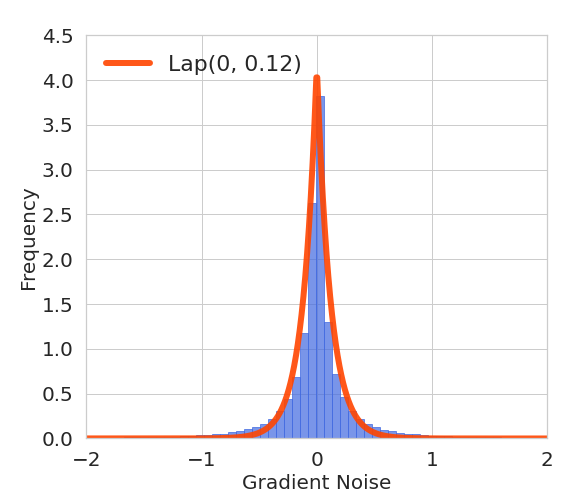}
        \subcaption{CIFAR-10, $80,000$ iterations, $L_2$ norm, $\rho=\frac{150}{255}$, Lap$(0,0.12)$.}
    \end{subfigure}
    \hspace{2mm}
    \begin{subfigure}[t]{0.23\linewidth}
        \includegraphics[width=\linewidth]{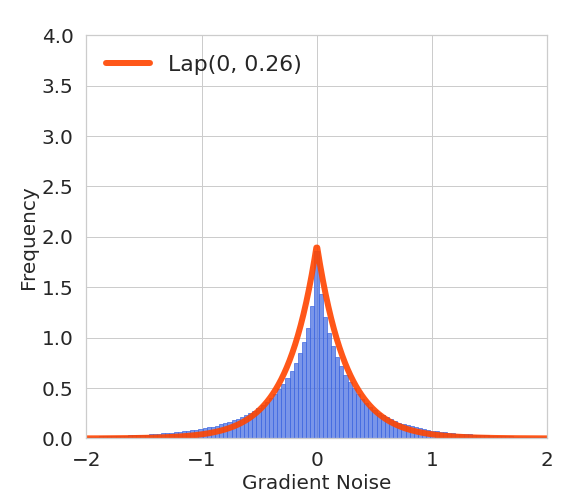}
        \subcaption{CIFAR-100, $80,000$ iterations, $L_\infty$ norm, $\rho=\frac{5}{255}$, Lap$(0,0.26)$.}
    \end{subfigure}
    \hspace{2mm}
    \begin{subfigure}[t]{0.23\linewidth}
        \includegraphics[width=\linewidth]{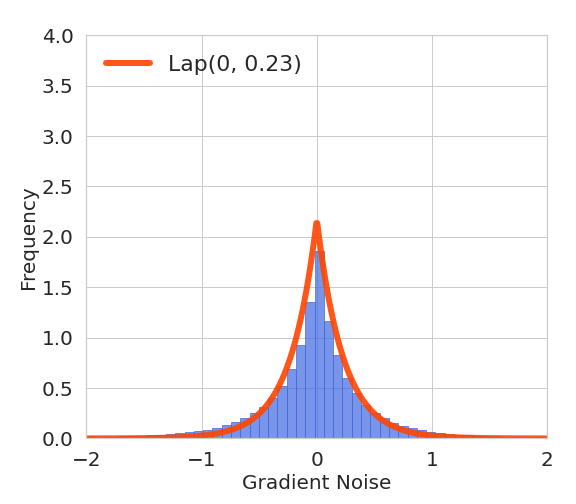}
        \subcaption{CIFAR-100, $80,000$ iterations, $L_2$ norm, $\rho=\frac{150}{255}$, Lap$(0,0.23)$.}
    \end{subfigure}

    \caption{Histograms of the gradient noises from different iterations in ERM under different experimental settings. Each plot is based on $10,000,000$ gradient noises.
    All the plots show that the distribution of gradient noise is similar to Laplacian distribution, which justifies that it is favorable to model gradient noise with Laplacian distribution.
    }
	\label{fig:lap}
\end{figure*}

\subsection{Is robustified intensity informative?}

A comprehensive empirical study is conducted, comparing empirical robustified intensity $\hat I_{1:T}$, radius $\rho$, and adversarial accuracy on the CIFAR-10 and CIFAR-100 datasets. Implementation details are given in Section \ref{sec:exp}. 

The empirical robustified intensity, radius, and adversarial accuracy are collected in every setting, as shown in fig. \ref{fig:robustness_intensity}. Fourth-order polynomial regression is employed for curve fitting.\footnote{We grid searched different values of the polynomial regression order in $\{1, 2, \ldots, 6 \}$. $4$ is the ``sweet point'' between ``under-fitting'' and ``over-fitting''. In the rest of this paper, we employ polynomial regression for curving fitting multiple times. The orders are selected similarly.}
From the collected data, we obtain two major observations:
(1) the (empirical) robustified intensity has a clear positive correlation with the radius $\rho$; 
and (2) the adversarial accuracy has a clear positive correlation with the robustified intensity is observed in the full interval of robustified intensity. The two observations verify that the robustified intensity is comparably informative to two standard measures for adversarial robustness, radius and adversarial accuracy.

\section{Privacy-robustness trade-off}
\label{sec:privacy_robustness}

This section studies the relationship between privacy preservation and robustness in adversarial training.
 
\subsection{What is the distribution of gradient noise?}

Stochastic gradient-based optimizers introduce noise in optimization.
It is interesting to ask {\it what is the distribution of gradient noise?} Some works \cite{kushner2003stochastic, ljung2012stochastic, mandt2017stochastic, jastrzkebski2017three, smith2018bayesian} assumed that the gradient noise is drawn from a Gauss distribution. In this work, we conducted a large-scale experiment to investigate the distribution of gradient noise.
The experiment results on CIFAR-10 and CIFAR-100 are collected in fig. \ref{fig:lap}, which demonstrate that Laplacian distribution is appropriate to model the distribution of gradient noise. Then, we make the following assumption. For more implementation details, please see Section \ref{sec:exp}.

\begin{assumption}
The gradient calculated from a mini-batch is drawn from a Laplacian distribution centered at the empirical risk,
\begin{align*}
\frac{1}{\tau} \sum_{(x, y) \in \mathcal B} \nabla_\theta \max_{\Vert x^\prime - x \Vert \leq \rho}\ell(h_\theta( x^\prime), y) \sim \mathrm{Lap} \left(\nabla_\theta \hat{\mathcal R}_S^A (\theta), b\right).
\end{align*}
\end{assumption}

\subsection{Theoretical evidence}
\label{sec:privacy_ri_theory}

Following the Laplacian mechanism, we approximate the differential privacy of adversarial training as the following two theorems.

\begin{theorem}[Privacy-robustness relationship I]
\label{thm:privacy_whole_prime}
Suppose one employs SGD for adversarial training and the whole training procedure has $T$ iterations.
Then, the adversarial training is $(\varepsilon_A, \delta_A)$-differentially private, where
\begin{gather}
	\nonumber
	\varepsilon_A = \sqrt{2 \log \frac{N}{\delta'} \sum_{t=1}^T \varepsilon^2_t } + \sum_{t=1}^T \varepsilon_t \frac{e^{\varepsilon_t} - 1}{e^{\varepsilon_t}+1}, \\
	\nonumber
	\delta_A =  \frac{\delta'}{N},
	\end{gather}
	in which
	\begin{gather*}
	\varepsilon_t = \frac{2 L^\mathrm{ERM}_t}{N b} I_t,
	\end{gather*}
and $\delta'$ is a positive real, $I_t$ is the single-iteration robustified intensity in the $t$-th iteration, and $b$ is the Laplacian parameter.
\end{theorem}

\begin{theorem}[Privacy-robustness relationship II]
\label{thm:privacy_whole}
Suppose a $T$-iteration SGD is employed to solve the minimax optimization problem in adversarial training. Then, the adversarial training is $(\varepsilon_A, \delta_A)$-differentially private, where
\begin{gather*}
	\varepsilon_A = \varepsilon_{1:T} \sqrt{2T \log \frac{N}{\delta^\prime}} + \mathcal{O}\left( \frac{1}{N^2} \right), \\
	\nonumber
	\delta_A =  \frac{\delta'}{N},
	\end{gather*}
	where
	\begin{gather*}
	\varepsilon_{1:T} = \frac{2 L^\mathrm{ERM}_{1:T}}{N b} I_{1:T},
	\end{gather*}
and $L^\mathrm{ERM}_{1:T} := \left(\frac{1}{T} \sum_{t=1}^T \left(L^\mathrm{ERM}_t\right)^4\right)^{\frac{1}{4}}$,
$\delta'$ is a positive real, $I_{1:T}$ is the robustified intensity for the whole training procedure (see Definition \ref{def:ri}), and $b$ is the Laplacian parameter.
\end{theorem}

\begin{remark}
Theorems \ref{thm:privacy_whole_prime} and \ref{thm:privacy_whole} suggest a negative correlation between adversarial robustness and privacy preservation. 
\end{remark}

\begin{remark}
Theorem \ref{thm:privacy_whole_prime} approximate the differential privacy of a learning algorithm based on the adversarial robustness of its every iteration ($I_1, \ldots, I_t$), while Theorem \ref{thm:privacy_whole} approximate the differential privacy based on the adversarial robustness of the whole algorithm ($I_{1:T}$).
\end{remark}

\begin{remark}
\label{remark:privacy_whole}
The approximation of differential privacy given by Theorem \ref{thm:privacy_whole} is $(\mathcal O (\sqrt{\log N}/N),\mathcal O (1/N))$.
\end{remark}


Similarly, we can approximate the differential privacy of ERM in the following corollary.
\begin{corollary}
\label{cor:privacy_iteration_ERM}
Suppose one employs SGD for ERM and  the whole training procedure has $T$ iterations. Then, the ERM is $(\varepsilon, \delta)$-differentially private, where
\begin{gather*}
		\varepsilon = \varepsilon^\mathrm{ERM}_{1:T} \sqrt{2 T \log \frac{N}{\delta'}} + \mathcal{O}\left( \frac{1}{N^2} \right),\\
	\delta =  \frac{\delta'}{N},
\end{gather*}
in which,
\begin{equation*}
\varepsilon^\mathrm{ERM}_{1:T} = \frac{2 L^\mathrm{ERM}_{1:T}}{N b},
\end{equation*}
and $L^\mathrm{ERM}_{1:T} := \left(\frac{1}{T} \sum_{t=1}^T \left(L^\mathrm{ERM}_t\right)^4\right)^{\frac{1}{4}}$,
$\delta^\prime$ is a positive real.
\end{corollary}

Comparing the results for adversarial training and ERM, we have
\begin{equation*}
\varepsilon_{1:T} = I_{1:T} \cdot \varepsilon^{\text{ERM}}_{1:T} + \mathcal{O}(1/N^2).
\end{equation*}
Theorem \ref{thm:privacy_whole} and Corollary \ref{cor:privacy_iteration_ERM} show that both factors $\varepsilon$ and $\delta$ have positive correlations with the robustified intensity, which suggests a trade-off between privacy preservation and adversarial robustness.

\subsubsection{Proof skeleton}

This section presents the proof skeleton. Detailed proofs for Theorems \ref{thm:privacy_whole_prime} and \ref{thm:privacy_whole} are given in Sections \ref{app:privacy} and \ref{proof:privacy_whole}, respectively. 

It is usually hard to calculate the differential privacy of iterative algorithm directly. However, the differential privacy of every single step in the iterative algorithm can be easily derived. Based on the differential privacy of every steps, composition theorems can then approximate the differential privacy of the whole learning algorithm. The proofs for calculating the differential privacy of adversarial training follows the intuition above.

In this paper, we employs the tightest composition theorem by He {\it et al.} \cite{he2020tighter} as follows.

\begin{lemma}[Advanced composition; cf. \cite{he2020tighter}, Theorem 5]
	\label{lemma:composition}
	Suppose an iterative algorithm has $T$ steps, and the $t$-th step is $\varepsilon_t$-differentially private. Then, the whole algorithm is $(\varepsilon, \delta)$-differentially private, where
	\begin{gather*}
	\varepsilon = \sqrt{2 \log \frac{1}{\delta} \sum_{t=1}^T \varepsilon^2_t} + \sum_{t=1}^T \varepsilon_t \frac{e^{\varepsilon_t} - 1}{e^{\varepsilon_t} + 1},
	\end{gather*}
	and $\delta$ is a positive real.
\end{lemma}

\begin{figure*}[t]
    \includegraphics[width=0.24\linewidth]{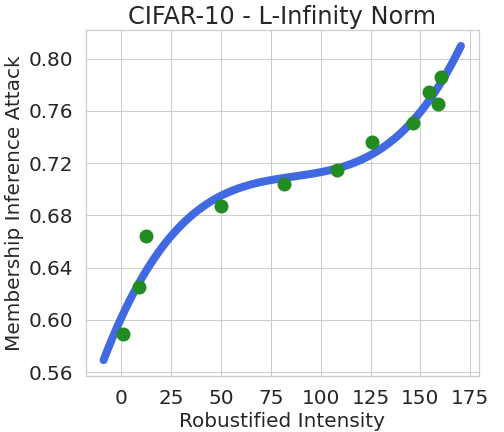}
    \includegraphics[width=0.24\linewidth]{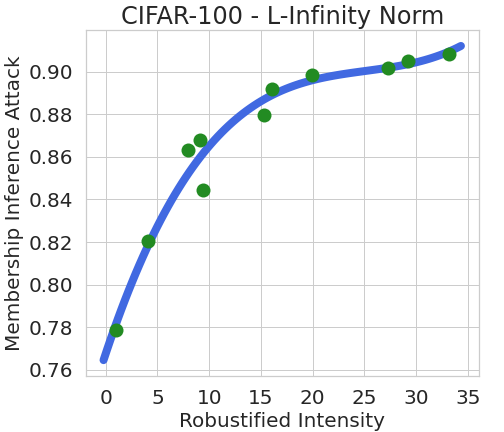}
    \includegraphics[width=0.24\linewidth]{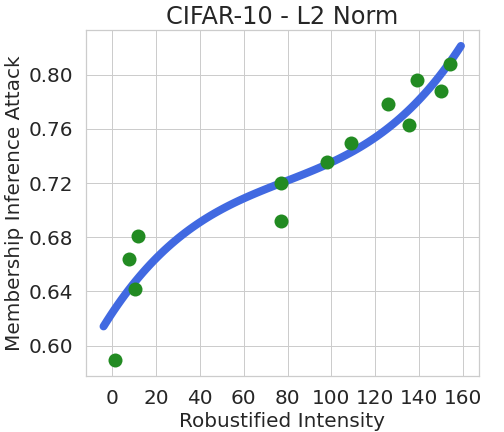}
    \includegraphics[width=0.24\linewidth]{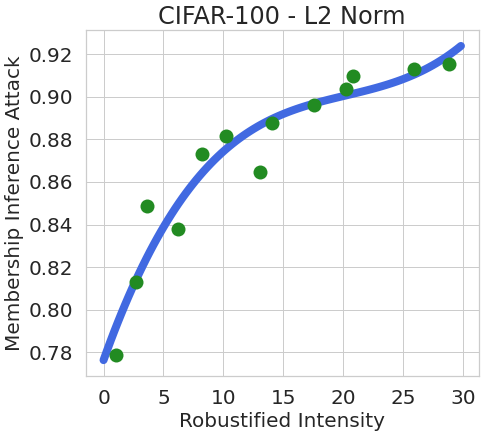}
\caption{Plots of membership inference attack accuracy vs. empirical robustified intensity. The datasets and adversarial training metric norms of the four plots are respectively (1) CIFAR-10 and $L_\infty$ norm; (2) CIFAR-100 and $L_\infty$ norm; (3) CIFAR-10 and $L_2$ norm; and (4) CIFAR-100 and $L_2$ norm. 
Fourth-order polynomial regression is used for curve fitting in each figure. Based on these figures, we find that membership inference attack accuracy has a positive correlation with robustified intensity, which demonstrates a negative correlation between privacy preservation and adversarial robustness.}

\label{fig:attack}
\end{figure*}

\subsection{Empirical evidence}


We trained Wide ResNet on datasets CIFAR-10 and CIFAR-100 to verify the privacy-robustness trade-off. For more implementation details, please see Section \ref{sec:exp}.

The privacy-preserving ability is measured by membership inference attack \cite{shokri2017membership, yeom2018privacy}, a standard privacy attack tool. Membership inference attack aims to inference whether a given data point comes from the training set based on the output of the model.
A higher membership inference attack accuracy means that a privacy attack for private information is more likely to succeed and thus implies a worse privacy-preserving ability.

The membership inference attack accuracies and empirical robustified intensities of all models are collected, as shown in fig. \ref{fig:attack}.
From the four figures, we observe a clear positive correlation between the membership inference attack accuracy and the robustified intensity $I_{1:T}$, which demonstrates a negative correlation between privacy preservation and robustness.

\section{Generalization-robustness trade-off}
\label{sec:generalization}

This section studies the relationship between generalization and robustness.

We first prove a high-probability generalization bound for an $(\varepsilon, \delta)$-differentially private machine learning algorithm. 
Combining the established privacy-robustness relationship, this bound helps study the generalizability of adversarial training.

\begin{theorem}[High-probability generalization bound via differential privacy]
\label{thm:high_probability_privacy}
Suppose all conditions of Theorem \ref{thm:privacy_whole} hold. Then, the algorithm $\mathcal A$ has a high-probability generalization bound as follows. Specifically, the following inequality holds with probability at least $1 - \gamma$:
\begin{align}
\label{eq:high_probability_privacy}
   & \mathbb{E}_{\mathcal{A}}\mathcal R(\mathcal{A}(S)) - \mathbb{E}_{\mathcal{A}}\hat{\mathcal R}_S(\mathcal{A}(S)) 
   \le c \left(M(1-e^{-\varepsilon}+e^{-\varepsilon}\delta) \log{N}\log{\frac{N}{\gamma}}+\sqrt{\frac{\log 1 / \gamma}{ N}} \right),
\end{align}
where $\gamma$ is an arbitrary probability mass, $M$ is the bound for loss $l$, $N$ is the training sample size, $c$ is a universal constant for any sample distribution,  and the probability is defined over the sample set $S$.
\end{theorem}

We also prove an on-average generalization bound, which expresses the ``expected'' generalizability. It worths noting that high-probability generalization bounds can also lead to on-average bounds by integration in theory. However, the calculations would be prohibitively difficult. Thus, we practice an independent approach to prove the on-average bound. 

 \begin{theorem}[On-average generalization bound via differential privacy]
\label{thm:uniform_generalization_privacy_2}
Suppose all conditions of Theorem \ref{thm:privacy_whole} hold. Then, the on-average generalization error of the algorithm $\mathcal A$ is upper bounded by
	\begin{equation*}
	\label{eq:uniform_generalization_privacy_2}
	\mathbb{E}_{S,\mathcal{A}}\left[\mathcal R(\mathcal{A}(S)) - \hat{\mathcal R}_S(\mathcal{A}(S))\right] \le M \delta e^{-\varepsilon} + M(1-e^{-\varepsilon}).
	\end{equation*}
\end{theorem}

\begin{remark}
By the Post-processing property of differential privacy, since $ \mathcal{B}$: $h\rightarrow \max_{x'\in \mathbb{B}_{*}(\rho)}\ell(h,(x',*))$ is a one-to-one mapping, $ \max_{x'\in \mathbb{B}_{*}(\rho)}\ell(\mathcal{A},(x',*))$ is $(\varepsilon,\delta)$ differentially private. Therefore, Theorem \ref{thm:high_probability_privacy} and \ref{thm:uniform_generalization_privacy_2} hold for adversarial learning algorithms.
\end{remark}

Both generalization bounds have positive correlations with the magnitude of the differential privacy, which further has a positive correlation with the adversarial robustness. This leads to the following corollary.

\begin{corollary}
There is a trade-off between generalizability and adversarial robustness (measured by robustified intensity) in adversarial training.
\end{corollary}


\subsection{Establishing generalization bounds based on algorithmic stability}

Theorems \ref{thm:high_probability_privacy} and \ref{thm:uniform_generalization_privacy_2} are established via algorithmic stability which measures how stable an algorithm is when the training sample is exposed to disturbance \cite{rogers1978finite, kearns1999algorithmic, bousquet2002stability}. While algorithmic stability has many different definitions, this paper mainly discusses the uniform stability.

\begin{definition}[Uniform stability; cf. \cite{bousquet2002stability}]
\label{def:uniform_stability}
A machine learning algorithm $\mathcal A$ is uniformly stable, if for any neighboring sample pair $S$ and $S'$ which are different by only one example, we have the following inequality,
\begin{equation*}
  \left\vert \mathbb E_{\mathcal{A}}\ell(\mathcal{A}(S), Z) - \mathbb E_{\mathcal{A}}\ell(\mathcal{A}(S'), Z) \right\vert \le \beta,
\end{equation*}
where $Z$ is an arbitrary example, $\mathcal{A}(S)$ and $\mathcal{A}(S')$ are the output hypotheses learned on the training sets $S$ and $S'$, respectively, and $\beta$ is a positive real constant. The constant $\beta$ is called the uniform stability of the algorithm $\mathcal A$.
\end{definition}

In this paper, we prove that $(\varepsilon, \delta)$-differentially private machine learning algorithms are algorithmic stable as the following lemma.

 \begin{lemma}[Stability-privacy relationship]
\label{lem:privacy_stability}
Suppose that a machine learning algorithm $\mathcal{A}$ is $(\varepsilon,\delta)$-differentially private. Assume the loss function $l$ is upper bounded by a positive real constant $M > 0$. Then, the algorithm $\mathcal A$ is uniformly stable,
\begin{equation*}
	\left\vert \mathbb E_{\mathcal{A}}\ell(\mathcal{A}(S), Z) - \mathbb E_{\mathcal{A}}\ell(\mathcal{A}(S'), Z) \right\vert \le M \delta e^{-\varepsilon} + M(1-e^{-\varepsilon}).
\end{equation*}
\end{lemma}


\subsubsection{Establishing high-probability generalization bound}

 \begin{figure*}[t]
    \includegraphics[width=0.24\columnwidth]{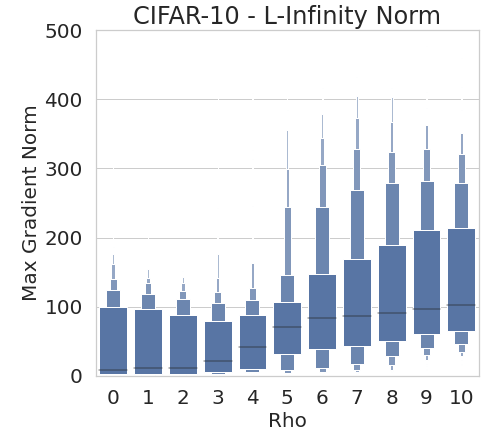}
    \includegraphics[width=0.24\columnwidth]{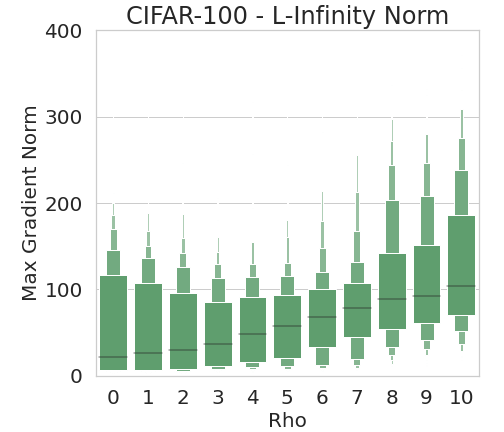}
    \includegraphics[width=0.24\columnwidth]{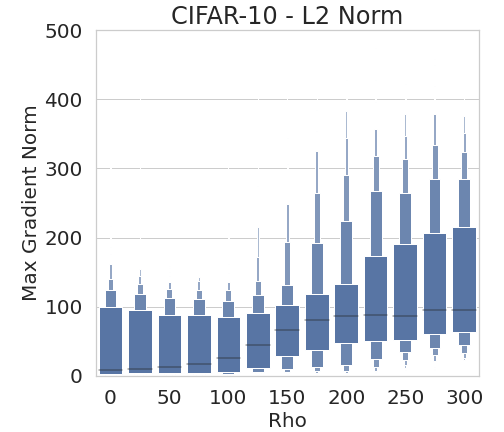}
    \includegraphics[width=0.24\columnwidth]{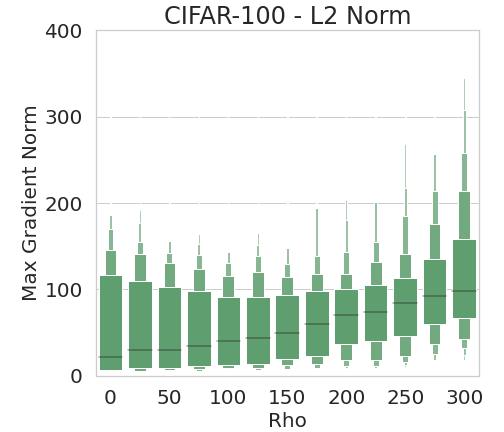}
    \caption{
    Box plots of the obtained $L^\mathrm{ERM}_t$ in ERM and $L^A_t$ in adversarial training on CIFAR-10 and CIFAR-100 with different radius $\rho$ and different adversarial training metric norms.
    In every setting, the training procedure has $800,000$ iterations, and the max gradient norms are calculated every $100$ iterations. Therefore, the sample sizes are $800$ in all settings.
    From the figures, we could find that the max gradient norms are small comparing to the sizes of hypothesis parameters, which verifies the tightness of our generalization bounds.
    }
    \label{fig:box}
\end{figure*}

Our high-probability generalization bound relies on the following lemma by Feldman {\it et al.} \cite{feldman2019high}.

\begin{lemma}[cf. \cite{feldman2019high}, Theorem 1]
	\label{lemma:generalization_stability_II}
	Suppose a deterministic machine learning algorithm $\mathcal A$ is stable with uniform stability $\beta$. Suppose $l\le 1$. Then, for any sample distribution and any $\gamma \in (0,1)$, there exists a universal constant $c$, such that, with probability at least $1-\gamma$ over the draw of sample, the generalization error can be upper bounded as follows,
	\begin{align*}
	& \mathbb{E}_{z\sim P} \ell(\mathcal{A}(S),z) - \frac{1}{N}\sum_{z\in S} \ell(\mathcal{A}(S),z) 
	\leq c \left(\beta \log{N}log{\frac{N}{\gamma}}+\sqrt{\frac{\log 1 / \gamma}{ N}}\right).
	\end{align*}
\end{lemma}

\begin{proof}[Proof of Theorem \ref{thm:high_probability_privacy}]
Combining Lemma \ref{lem:privacy_stability} and Lemma \ref{lemma:generalization_stability_II}, we can directly prove Theorem \ref{thm:high_probability_privacy}.
\end{proof}

\subsubsection{Establishing on-average generalization bound}

Our on-average generalization bound relies on the following lemma by Dwork {\it et al.} \cite{dwork2015generalization}.

\begin{lemma}[Lemma 11, cf. \cite{shalev2010learnability}]
\label{lemma:on_average_generalization_general}
Suppose the loss function is upper bounded. For any machine learning algorithm with $\beta$ Replace-one stability, its generalization error is upper bound as follows,
\begin{equation*}
\mathcal R(\mathcal{A}(S)) - \hat{\mathcal R}_S(\mathcal{A}(S)) \le \beta.
\end{equation*}
\end{lemma}

\begin{proof}[Proof of Theorem \ref{thm:uniform_generalization_privacy_2}]
Combining Lemma \ref{lem:privacy_stability} and Lemma \ref{lemma:on_average_generalization_general}, we can directly prove Theorem \ref{thm:uniform_generalization_privacy_2}.
\end{proof}

\subsection{Tightness of generalization bounds}
\label{sec:tightness}

This section analyses the tightness of generalization bounds.

\textbf{Dependency on the training sample size $N$.} In Section \ref{sec:privacy_robustness}, we have approximated the rate of adversarial training's differential privacy with respect to the training sample size $N$; see Remark \ref{remark:privacy_whole}. Combining Theorems \ref{thm:high_probability_privacy} and \ref{thm:uniform_generalization_privacy_2}, we can approximate the tightness of the high-probability generalization bound and the on-average generalization bound as the following two remarks.

\begin{remark}
\label{remark:high_probability_generalization}
The high-probability generalization bound given by Theorem \ref{thm:high_probability_privacy} is $\mathcal O (1/\sqrt N)$.
\end{remark}

\begin{remark}
\label{remark:on_average_generalization}
The on-average generalization bound given by Theorem \ref{thm:uniform_generalization_privacy_2} is $\mathcal O (\sqrt{\log N}/N)$.
\end{remark}

\textbf{Dependency on the model size.} Our generalization bounds do not explicitly rely on the parameter size, which would be prohibitively large in deep learning.
The only terms that could rely on the model size are the max gradient norms. 
We empirically investigated their magnitude. 
We trained Wide ResNets on CIFAR-10, CIFAR-100, under the frameworks of ERM and adversarial training with multiple different values of the radius $\rho$ and two different metric norms in adversarial training, $L_2$ norm and $L_\infty$ norm.
The collected data is shown in fig. \ref{fig:box}. The box plots clearly demonstrate that the gradient norms are very small comparing to the parameter size, which would be tens of millions.

\textbf{Comparing with existing bounds.} 
Existing works have proved generalization bounds based on hypothesis complexity. Yin et al. \cite{yin2019rademacher} prove an $\mathcal O(1/\sqrt{N})$ generalization bound for models learned by adversarial training, based on Rademacher complexity of the deep neural networks. Khim and Loh \cite{khim2018adversarial} also prove an $\mathcal O(1/\sqrt{N})$ generalization bound based on the Rademacher complexity of the hypothesis space. Tu et al. \cite{tu2019theoretical} prove an $\mathcal O(1/\sqrt{N})$ generalization bound based on the covering number of the hypothesis space. All these bounds heavily rely on the hypothesis complexity, which would, however, be prohibitively large in deep learning.

\begin{figure*}[t]
    \includegraphics[width=0.24\linewidth]{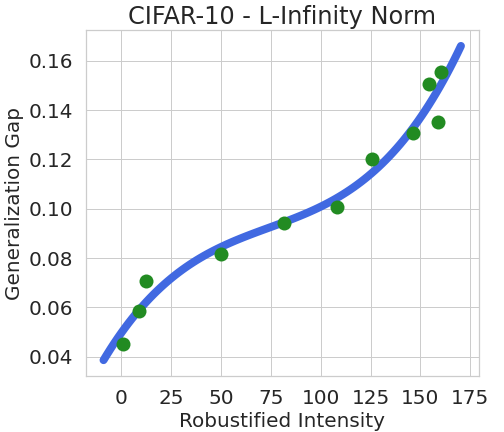}
    \includegraphics[width=0.24\linewidth]{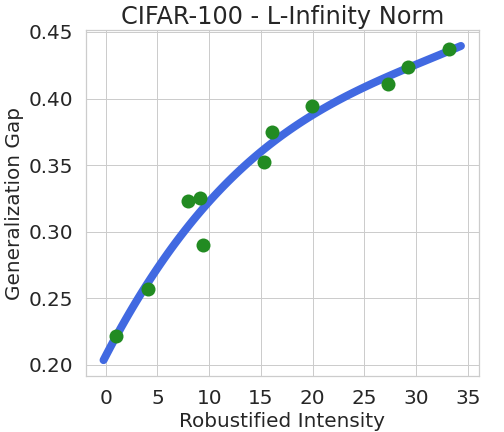}
    \includegraphics[width=0.24\linewidth]{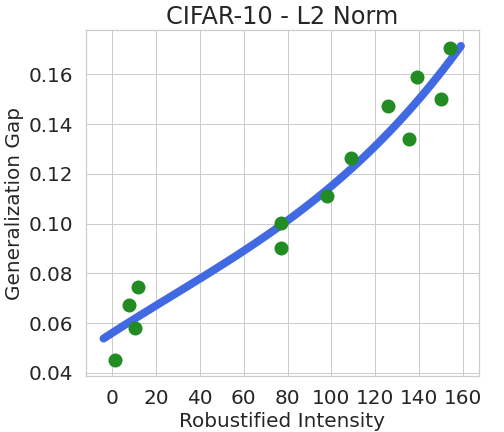}
    \includegraphics[width=0.24\linewidth]{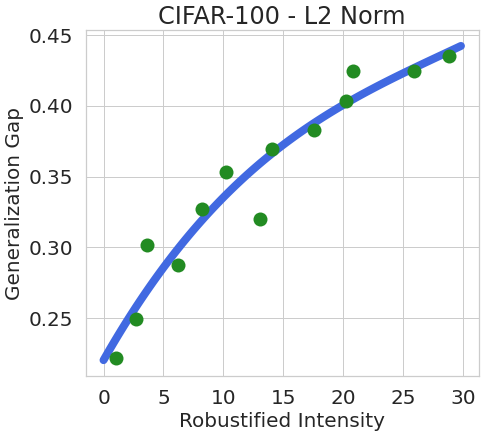}
\caption{Plots of generalization gap vs. empirical robustified intensity. The datasets and PGD metric norms of the four plots are respectively (1) CIFAR-10 and $L_\infty$ norm; (2) CIFAR-100 and $L_\infty$ norm; (3) CIFAR-10 and $L_2$ norm; and (4) CIFAR-100 and $L_2$ norm.
Fourth-order polynomial regression is used for curve fitting in each figure. Based on these figures, we find that generalization gap has a positive correlation with robustified intensity, which demonstrates a negative correlation between privacy preservation and adversarial robustness.
}
\label{fig:generalization_gap}
\end{figure*}

\subsection{Experimental evidence}

The empirical study on the generalization-robustness trade-off is based on Wide ResNet and datasets of CIFAR-10, CIFAR-100. The generalization abilities of all models are evaluated by the generalization gap, which is defined to be the difference between the training accuracy and test accuracy. 
For more implementation details, please see Section \ref{sec:exp}.

We collect the generalization gaps and empirical robustified intensities of all models. Based on the collected data, four plots are shown in fig. \ref{fig:generalization_gap}.
Fourth-order polynomial regression is employed for curve fitting.
A larger generalization gap implies a worse generalizability. From the plots, one major observation is obtained: the generalization gap has a clear positive correlation with the robustified intensity, which verifies the generalization-robustness trade-off.

\section{Proofs}
\label{app:proofs}
In this section, we provide detail proofs omitted in the previous sections.

\subsection{Proof of Theorem \ref{thm:asymtotic}}
\label{app:asymtotic}

We first recall additional preliminaries that necessary in the proofs.

Suppose every example $z$ is independently and identically (i.i.d.) sampled from the data distribution is $\mathcal{D}$; {\it i.e.}, $z \sim \mathcal D$. Thus, the training sample set $S \sim {\mathcal D}^N$, where $N$ is the training sample size. 

Besides, we need the following two definitions in the rest of the paper.

\begin{definition}[Ball and sphere]
The ball in space $\mathcal H$ centered at point $x \in \mathcal H$ of radius $r > 0$ in term of norm $\| \cdot \|$ is denoted by 
\begin{equation*}
\mathbb{B}_{h}(r) = \{x: \| x - h \| \le r \}.
\end{equation*}
The sphere $\partial\mathbb{B}_{h}(r)$ of ball $\mathbb{B}_{h}(r)$ is defined as below,
\begin{equation*}
\partial\mathbb{B}_{h}(r) = \{x: \| x - h \| = r \}.
\end{equation*}
\end{definition}

\begin{definition}[Complementary set]
For a subset $A \subset \mathcal H$ of a space $\mathcal H$, its complementary set $A^c$ is defined as below,
\begin{equation*}
 A'=\{h:h\in \mathcal{H}, h\not\in A\}.
\end{equation*}
\end{definition}

\begin{proof}[Proof of Theorem \ref{thm:asymtotic}]
		We only need to prove that almost surely
		\begin{align}
		& \lim_{\tau\rightarrow\infty}\max _{(x_{i}, y_{i}) \in \mathcal{B}}\left\|\nabla_{\theta} \max _{\left\|x_{i}^{\prime}-x_{i}\right\| \leq \rho} l\left(h_{\theta}\left(x_{i}^{\prime}\right), y_{i}\right)\right\| 
		 \quad =\max _{x, y}\left\|\nabla_{\theta} \max _{\left\|x^{\prime}-x\right\| \leq \rho} l\left(h_{\theta}\left(x^{\prime}\right), y\right)\right\|, \label{eq:limits_max}
		\end{align}
		and almost surely
		\begin{align}
		&\lim_{\tau\rightarrow\infty}\max _{(x_{i}, y_{i}) \in \mathcal{B}}\left\|\nabla_{\theta} l\left(h_{\theta}\left(x_{i}\right), y_{i}\right)\right\| 
		\quad =\max _{x, y}\left\|\nabla_{\theta} l\left(h_{\theta}\left(x\right), y\right)\right\|. \label{eq:limits_original}
		\end{align}
		
		We first prove that for any positive real $\rho > 0$, \begin{equation}
		g(\theta,z)=\nabla_{\theta} \max _{x'\in \mathbb{B}_{x}(\rho)} l\left(h_{\theta}\left(x^{\prime}\right), y\right)
		\end{equation}
		is a continuous function with respect to $z=(x,y)$, where
		\begin{equation*}
		\mathbb{B}_{x}(\rho) = \{x': \| x - x' \| \le \rho \}
		\end{equation*}
		is a ball centered at $x$ with radius of $\rho$. 
		
		 Fixing $y\in \mathcal{Y}$, define
		 \begin{equation*}
		 T_y(x)=\arg\max_{x'\in \mathbb{B}_{x}(\rho)}\ell(h_\theta(x'),y)
		 \end{equation*}
		 as a mapping from $\mathcal{X}$ to $\mathcal{X}$. We will prove $T_y(x)$ is continuous with respect to $(x,y)$ by reduction to absurdity. Suppose there exists a sequence
		 \begin{equation*}
		     \{z_i=(x^i,y^i)\}_{i=1}^\infty, \text{ } \lim\limits_{i\rightarrow\infty} z_i=z_0,
		 \end{equation*}
		 and a constant $\varepsilon_A>0$ such that
		 \begin{equation*}
		     \Vert T_{y^i}(x^i)-T_{y^0}(x^0)\Vert\ge \varepsilon_A.
		 \end{equation*}
		 
		 Since $\{T_{y^i}(x^i)\}_{i=1}^\infty$ is a bounded set, there exists an increasing subsequence $\{k_i\}_{i=1}^\infty\subseteq Z^+$ such that $\{T_{y_{k_i}}(x_{k_i})\}_{i=1}^\infty$ converges to some point $T_\infty$. Then, we have that 
		\begin{equation*}
 		T_\infty\in \cup_{i=1}^{\infty}\cap_{j=i}^{\infty} \mathbb{B}_{x_{k_i}}(\rho)\subset\mathbb{B}_{x^0}(\rho).
		\end{equation*} 
		
		Furthermore, for any $\varepsilon\ge 0$, there exists a $\delta > 0$, such that for any $x\in \mathbb{B}_{T_{y^0}(x^0)}(\delta)$,  
		\begin{equation*}
		\ell(h_\theta(x),y^0))\ge \ell(h_\theta(T_{y^0}(x^0)),y^0)-\varepsilon.
		\end{equation*}
		{In case $T_{y^0}(x^0)\in \partial\mathbb{B}_{x^0}(\rho)$ such that $T_{y^0}(x^0)\not\subset\cap_{i=1}^{\infty }\mathbb{B}_{x_{k_i}}(\rho)$, } let $x'\in \mathbb{B}_{T_{y^0}(x^0)}(\delta)$ be an inner point of $\mathbb{B}_{x^0}(\rho)$. When $i$ is large enough, we have $x'\in \mathbb{B}_{x_{k_i}}(\rho)$, which yields 
		\begin{equation*}
		\ell(h_{\theta}(x'),y_{k_i})\le 	\ell(h_{\theta}(T_{y_{k_i}}(x_{k_i})),y_{k_i}).
		\end{equation*}
		Let $i$ approaches $\infty$, we then have
		\begin{equation*}
		\ell(h_\theta(T_{y^0}(x^0),y^0)-\varepsilon\le \ell(h_{\theta}(x'),y^0)\le 	\ell(h_{\theta}(T_\infty),y^0).
		\end{equation*}
		Since $\varepsilon$ is arbitrarily selected, we then have
		\begin{equation*}
		\ell(h_\theta(T_{y^0}(x^0),y^0))\le 	\ell(h_{\theta}(T_\infty),y^0)\le \ell(h_\theta(T_{y^0}(x^0),y^0)).
		\end{equation*}
		Therefore, $T_\infty=T_{y^0}(x^0)$, which leads to a contradictary since
		\begin{equation*}
		\Vert T_{y^i}(x^i)-T_{y^0}(x^0)\Vert\ge \varepsilon_A.
		\end{equation*}
		
		Since $g(\theta,z)$ can be rewritten as 
		\begin{equation*}
		g(\theta,z)=\nabla_{\theta} \max _{x'\in \mathbb{B}_{x}(\rho)} l\left(h_{\theta}\left(x^{\prime}\right), y\right)=\nabla_{\theta} \ell(h_{\theta}(T_{y}(x)),y),
		\end{equation*}
		by Assumption \ref{assum:continuity}, we have $	g(\theta,z)$ is continuous with respect to $z$.
	
		Now we can prove eq. (\ref{eq:limits_max}) and eq. (\ref{eq:limits_original}). As for eq. (\ref{eq:limits_max}), there exist $z_0=(x^0, y^0)$ such that
		\begin{align*}
	&\left\|\nabla_{\theta} \max _{\left\|x^{\prime}-x^0\right\| \leq \rho} l\left(h_{\theta}\left(x^{\prime}\right), y^0\right)\right\| 
	\quad =\max _{(x, y)}\left\|\nabla_{\theta} \max _{\left\|x^{\prime}-x\right\| \leq \rho} l\left(h_{\theta}\left(x^{\prime}\right), y\right)\right\|.
		\end{align*}
		
	    For any $\varepsilon > 0$, since $g(\theta,z)$ is continuous with respect to $z$, there exists a $\delta > 0$, such that for any $(x',y')\in \mathbb{B}_{(x^0,y^0)}(\delta)$,  
	    \begin{equation*}
	    g(\theta,(x',y'))\ge g(\theta,(x^0,y^0))-\varepsilon.
	     \end{equation*}
	     
	    Therefore,
	    \begin{equation*}
	    \{(x,y):g(\theta,(x,y))< g(\theta,(x^0,y^0))-\varepsilon\} \subset \left(\mathbb{B}_{(x^0,y^0)}(\delta)\right)^c,
	    \end{equation*}
	    and we have that
	    \begin{align*}
	    &\mathbb{P}_{\mathcal{B}\sim \mathcal{D}^\tau} \left(\max_{ z_i \in \mathcal{B}} g(\theta,z_i) < \max_{ z} g(\theta,z)-\varepsilon\right)
	    \\\le& \mathbb{P}_{\mathcal{B}\sim \mathcal{D}^\tau} \left(\max_{z_i \in \mathcal{B}} g(\theta,z_i) <  g(\theta,z_0)-\varepsilon\right)
	    \\
	    \le& \mathbb{P}_{\mathcal{B}\sim \mathcal{D}^\tau} \left(\mathcal{B}\cap \mathbb{B}_{(x^0,y^0)}(\delta)=\emptyset \right)
	    \\
	    =&\left(1-\mathbb{P}_{z\sim \mathcal{D}} \left(z\in\mathbb{B}_{(x^0,y^0)}(\delta)  \right)\right)^\tau.
	    \end{align*}
	    As $\tau\rightarrow\infty$, we have 
	    \begin{equation*}
	      \lim_{\tau \to \infty}\mathbb{P}_{\mathcal{B}\sim \mathcal{D}^\tau} \left(\max_{z_i \in \mathcal{B}} g(\theta,z_i)\le \max_{z} g(\theta,z)-\varepsilon\right)=0.
	    \end{equation*}
	    Since $\varepsilon$ is arbitrarily selected, we have
	    \begin{equation*}
	    \lim_{\tau \to \infty}\mathbb{P}_{\mathcal{B}\sim \mathcal{D}^\tau} \left(\max_{z_i \in \mathcal{B}} g(\theta,z_i)\le \max_{z} g(\theta,z)\right)=0,
	   \end{equation*}
	   which proves eq. (\ref{eq:limits_max}).
	   
	   Replacing $g(\theta,z)=\nabla_{\theta}\ell(h_{\theta}(x),y)$, we can prove  eq. (\ref{eq:limits_original}) following the same routine.
	   
	   The proof is completed.
\end{proof}

\subsection{Proof of Theorem \ref{thm:privacy_whole_prime}}
\label{app:privacy}

This section proves Theorem \ref{thm:privacy_whole_prime}. We first prove two lemmas.

Practically, high-probability approximations of $\varepsilon$-differential privacy are easier to be obtained from concentration inequalities. Lemma \ref{lemma:sample_dp} presents a relationship from high-probability approximations of $\varepsilon$-differential privacy to approximations of $(\varepsilon, \delta)$-differential privacy. Similar arguments are used in some related works; see, for example, the proof of Theorem 3.20 in \cite{dwork2014algorithmic}. Here, we give a detailed proof to make this paper completed.

\begin{lemma}
\label{lemma:sample_dp}
Suppose $\mathcal A:\mathcal{Z}^N\rightarrow \mathcal{H}$ is a stochastic algorithm, whose output hypothesis learned on the training sample set $S$ is $\mathcal A(S)$. For any hypothesis  $h\in\mathcal{H}$, if for probability at least $1 - \delta$
over the randomness of $\mathcal{A}(S)$ ,
\begin{equation}
\label{eq:sample_dp}
\log \left[ \frac{\mathbb P \left[ \mathcal A(S) =h \right]}{\mathbb P \left[ \mathcal A(S') = h\right]} \right] \le \varepsilon,
\end{equation}
the algorithm $\mathcal A$ is $(\varepsilon, \delta)$-differentially private.
\end{lemma}

\begin{proof}
	
	After rearranging eq. (\ref{eq:sample_dp}), we have that for probability at least $1 - \delta$,
	\begin{equation}
	\label{eq:sample_dp_rearranged}
	\mathbb P \left[ \mathcal A(S) = h \right] \le \mathbb P \left[ \mathcal A(S') = h \right] e^\varepsilon,
	\end{equation}
	
	For the brevity, we define an event as follows,
	\begin{equation*}
	B_0 = \left \{ h: \log \left[ \frac{\mathbb P \left[ \mathcal A(S) = h \right]}{\mathbb P \left[ \mathcal A(S') = h \right]} \right] \le \varepsilon \right \}.
	\end{equation*}
	
	Also, define that
	\begin{equation*}
	B_0^c = \mathcal{H} - B_0.
	\end{equation*}
	Apparently, for any subset $B \in \mathcal{H}$,
	\begin{align}
	\label{eq:algorithm_inequality}
	\mathbb P \left[ \mathcal A(S) \in B_0 \cap B \right] \le & \mathbb P \left[ \mathcal A(S') \in B_0 \cap B \right] e^\varepsilon,\\
	\mathbb P \left[ \mathcal A(S) \in B_0 \right] \ge & 1 - \delta,\nonumber\\
	\label{eq:B_c_inequality}
	\mathbb P \left[ \mathcal A(S) \in B_0^c \right] \le & \delta.
	\end{align}
	
	Then, for any subset $B \in \Theta$, we have that
	\begin{align*}
	& \mathbb P \left[ \mathcal A(S) \in B \right] \nonumber\\
	= & \mathbb P \left[ \mathcal A(S) \in B \cap \left( B_0 \cup B_0^c \right) \right] \nonumber\\
	= & \mathbb P \left[ \mathcal A(S) \in B \cap B_0 \right] + \mathbb P \left[ \mathcal A(S) \in B \cap B_0^c \right] \nonumber\\
	\le & \mathbb P \left[ \mathcal A(S) \in B \cap B_0 \right] + \mathbb P \left[ \mathcal A(S) \in B_0^c \right].
	\end{align*}
	
	Combining eqs. (\ref{eq:sample_dp_rearranged}), (\ref{eq:algorithm_inequality}), and (\ref{eq:B_c_inequality}), we have that
	\begin{align*}
	&\mathbb P \left[ \mathcal A(S) \in B \right] \\
	\le& e^\varepsilon \mathbb P \left[ \mathcal A(S') \in B \cap B_0 \right] + \delta \\
	\le& e^\varepsilon \mathbb P \left[ \mathcal A(S') \in B \right] + \delta.
	\end{align*}
	
	Therefore, the stochastic algorithm $\mathcal A$ is $(\varepsilon, \delta)$-differentially private.
	
	The proof is completed.
\end{proof}

We now prove the Theorem \ref{thm:privacy_whole_prime}.

\begin{proof}[Proof of Theorem \ref{thm:privacy_whole_prime}]
We assume that the gradients calculated from random sampled mini batch $\mathcal{B}$ with size $\tau$ are random variables drawn from a Laplace distribution (see a justification in the main text):
\begin{align*}
\frac{1}{\tau}\sum_{z\in \mathcal{B}}\nabla_\theta \max_{\Vert x^\prime - x \Vert \leq \rho}\ell(\theta, x, y) \sim \mathrm{Lap} \left(\nabla_\theta \hat{\mathcal R}_S^A (\theta), b\right).
\end{align*}
Correspondingly, its counterpart on the training sample set $S'$ is as follows,
\begin{align*}
\frac{1}{\tau}\sum_{z\in \mathcal{B}'}\nabla_\theta \max_{\Vert x^\prime - x \Vert \leq \rho}\ell(\theta, x, y) \sim \mathrm{Lap} \left(\nabla_\theta \hat{\mathcal R}_{S'}^A (\theta), b\right),
\end{align*}
where $\mathcal{B}'$ is uniformly sampled from $S'$ with size $\tau$.

The output hypothesis is uniquely indexed by the weight. Specifically, we denote the weight after the $t$-th iteration as $\theta^A_{t + 1}$. Furthermore, the weight updates $\Delta \theta^A_t = \theta^A_{t + 1} - \theta^A_t$ are uniquely determined by the gradients. Therefore, we can calculate the probability of the gradients to approximate the differential privacy. For any $\hat g_t^A$,
\begin{align}
\label{eq:pr_ratio}
& \log \left[ \frac{p \left[ \mathrm{Lap}(\nabla_{\theta} \hat{\mathcal R}^A_S (\theta^A_t), b) = \hat g^A_t \right]}{ p \left[ \mathrm{Lap}(\nabla_{\theta} \hat{\mathcal R}^A_{S'} (\theta^A_t), b) = \hat g^A_t  \right]} \right] \nonumber\\
= & \log \left[ \frac{\exp \left\{ - \left \| \nabla_{\theta} \hat{\mathcal R}^A_S(\theta^A_t) - \hat g^A_t  \right \| /  b \right\}}{ \exp \left\{ - \left \| \nabla_{\theta} \hat{\mathcal R}^A_{S'} (\theta^A_t) - \hat g^A_t  \right \|  /  b \right\}} \right] \nonumber\\
= & \frac{1}{ b} \left [ - \left \| \nabla_{\theta} \hat{\mathcal R}^A_S (\theta^A_t) - \hat g^A_t  \right \| + \left \| \nabla_{\theta} \hat{\mathcal R}^A_{S'} (\theta^A_t) - \hat g^A_t  \right \| \right].
\end{align}

Define that
\begin{equation*}
L^A_t = \max_{(x,y)} \left \| \nabla_{\theta} \max_{\Vert x^\prime - x \Vert \leq \rho}\ell(\theta^A_t, x^\prime, y) \right \|,
\end{equation*}
and
\begin{equation*}
v = \nabla_{\theta} \hat{\mathcal R}^A_{S'} (\theta^A_t) - \nabla_{\theta} \hat{\mathcal R}^A_{S} (\theta^A_t).
\end{equation*}
Because there only one pair of examples is different between the training sample set pair $S$ and $S'$, we have that,
\begin{gather}
\label{eq:v_norm}
\| v \| \le \frac{2 L_A}{N}.
\end{gather}

Combining eqs. (\ref{eq:pr_ratio}) and (\ref{eq:v_norm}), we have that
\begin{align}
\label{eq:pr_ratio_prime}
& \log \left[ \frac{p \left[ \mathrm{Lap}(\nabla_{\theta} \hat{\mathcal R}^A_S (\theta^A_t), b) = \hat g^A_t \right]}{ p \left[ \mathrm{Lap}(\nabla_{\theta} \hat{\mathcal R}^A_{S'} (\theta^A_t), b) = \hat g^A_t  \right]} \right] \nonumber\\
= &\frac{1}{ b} \left [ - \left \| \nabla_{\theta} \hat{\mathcal R}^A_S (\theta^A_t) - \hat g^A_t  \right \| + \left \| \nabla_{\theta} \hat{\mathcal R}^A_{S'} (\theta^A_t) - \hat g^A_t  \right \| \right]\nonumber\\
= & \frac{2L_A}{Nb}.
\end{align}

Since $L^A_t = I_t L^\mathrm{ERM}_t$, we have that
\begin{equation*}
 \log \left[ \frac{p \left[ \mathrm{Lap}(\nabla_{\theta} \hat{\mathcal R}^A_S (\theta^A_t), b) = \hat g^A_t \right]}{ p \left[ \mathrm{Lap}(\nabla_{\theta} \hat{\mathcal R}^A_{S'} (\theta^A_t), b) = \hat g^A_t  \right]} \right] \le \frac{2L^\mathrm{ERM}_t}{Nb} I_t.
\end{equation*}

Define that
\begin{equation*}
\varepsilon_t = \frac{2L^\mathrm{ERM}_t}{Nb} I_t.
\end{equation*}

 Applying Lemma \ref{lemma:composition} with $\varepsilon_1, \cdots, \varepsilon_T$ and $\delta=\frac{\delta'}{N}$, the proof is completed. 
\end{proof}

\subsection{Proof of Theorem \ref{thm:privacy_whole}}
\label{proof:privacy_whole}

\begin{proof}[Proof of Theorem \ref{thm:privacy_whole}]
By Theorem \ref{thm:privacy_whole_prime}, we have that the adversarial training is $(\varepsilon_A,\delta_A)$-differentially private, where
\begin{gather*}
	\varepsilon_A = \sqrt{2 \log \frac{N}{\delta'} \sum_{t=1}^T \varepsilon^2_t } + \sum_{t=1}^T \varepsilon_t \frac{e^{\varepsilon_t} - 1}{e^{\varepsilon_t}+1},
\end{gather*}
and $\delta_A = \frac{\delta^\prime}{N}$, $\varepsilon_t = \frac{2 L^\mathrm{ERM}_t}{N b} I_t$.

To bridge the gap between differential privacy and the robustified intensity $I_{1:T}$, we then bound $\varepsilon_A$ as follows,
\begin{align}
	\varepsilon_A
	&\leq \sqrt{2 \log \frac{N}{\delta'} \sum_{t=1}^T \varepsilon^2_t } + \frac{1}{2} \sum_{t=1}^T \varepsilon_t (e^{\varepsilon_t} - 1) \nonumber \\
	&= \sqrt{2 \log \frac{N}{\delta'} \sum_{t=1}^T \varepsilon^2_t } + \mathcal{O}\left(\frac{1}{N^2}\right). \label{eq:thm:privacy_whole:eps0_upper_bound}
\end{align}
Notice that the first term in eq. (\ref{eq:thm:privacy_whole:eps0_upper_bound}) is $\mathcal{O}(\sqrt{\log N}/N)$, which suggests that eq. (\ref{eq:thm:privacy_whole:eps0_upper_bound}) is dominated by its first term and the factor $\sum_{t=1}^T \varepsilon^2_t$ is a good indicator to measure the privacy preserving ability of the adversarial training. Therefore, we further bound the factor $\sum_{t=1}^T \varepsilon^2_t$ as follows,
\begin{align}
	\sum_{t=1}^T \varepsilon^2_t
	&= \sum_{t=1}^T \left( \frac{2 L^{ERM}_t}{N b} I_t \right)^2 \nonumber \\
	&\leq \sqrt{\sum_{t=1}^T \left(\frac{2L^{ERM}_t}{Nb}\right)^4} \cdot \sqrt{\sum_{t=1}^T I_t^4} \label{eq:thm:privacy_whole:ineq_cauchy} \\
	&= \sqrt{\left(\frac{2}{Nb}\right)^4 \cdot T \cdot \left(L^\mathrm{ERM}_{1:T}\right)^4} \cdot \sqrt{T \cdot I^4_{1:T}} \nonumber \\
	&= T \cdot \left(\frac{2 L^\mathrm{ERM}_{1:T}}{Nb} I_{1:T} \right)^2
	= T \cdot \varepsilon^2_{1:T}, \label{eq:thm:privacy_whole:sum_eps_upp_bd}
\end{align}
where eq. (\ref{eq:thm:privacy_whole:ineq_cauchy}) follows by Cauchy–Schwarz inequality. Inserting eq. (\ref{eq:thm:privacy_whole:sum_eps_upp_bd}) into eq. (\ref{eq:thm:privacy_whole:eps0_upper_bound}), we have that
\begin{gather*}
	\varepsilon_A \leq \varepsilon_{1:T} \sqrt{2T \log \frac{N}{\delta^\prime}} + \mathrm{O}\left( \frac{1}{N^2} \right).
\end{gather*}

Finally, defining
$\varepsilon := \varepsilon_{1:T} \sqrt{2T \log \frac{N}{\delta^\prime}} + \mathcal{O}\left( \frac{1}{N^2} \right)$ and
$\delta := \delta_A = \frac{\delta^\prime}{N}$,
then according to the definition of differential privacy, we can conclude that the adversarial training is also $(\varepsilon,\delta)$-differentially private, which completes the proof.
\end{proof}



\subsection{Proof of Lemma \ref{lem:privacy_stability}}
This section proves the relationship between differential privacy and uniform stability.

We first prove a weaker version of Lemma \ref{lem:privacy_stability} when algorithm $\mathcal{A}$ has $\varepsilon$-pure differential privacy.

\begin{lemma}
	Suppose a machine learning algorithm $\mathcal{A}$ is $\varepsilon$-differentially private. Assume the loss function $l$ is upper bounded by a positive real constant $M > 0$. Then, the algorithm $\mathcal A$ is uniformly stable,
	\begin{equation*}
	\left\vert \mathbb E_{\mathcal{A}(S)}\ell(\mathcal{A}(S), Z) - \mathbb E_{\mathcal{A}(S')}\ell(\mathcal{A}(S'), Z) \right\vert \le M(1-e^{-\varepsilon}).
	\end{equation*}
\end{lemma} 

\begin{proof}
	Let set $B$ defined as $B=\{h\in H: \ell(h,z)>t\}$, where $t$ is an arbitrary real. Then, for any $t \in \mathbb R$,
	\begin{equation}\label{eq:transdp}
	\mathbb P_{\mathcal{A}(S)}(\mathcal{A}(S)\in B)\le e^{\varepsilon} \mathbb P_{\mathcal{A}(S')}(\mathcal{A}(S')\in B).
	\end{equation}
	
	By rearranging eq. (\ref{eq:transdp}), we have
	\begin{gather}
	\nonumber e^{-\varepsilon}\mathbb P_{\mathcal{A}(S)}(\mathcal{A}(S)\in B) \le \mathbb P_{\mathcal{A}(S')}(\mathcal{A}(S')\in B),
	\end{gather}
	and
	\begin{align*}
	&(e^{-\varepsilon}-1)\mathbb P_{\mathcal{A}(S)}(\mathcal{A}(S)\in B) 
	\le \mathbb P_{\mathcal{A}(S')}(\mathcal{A}(S')\in B)-\mathbb P_{\mathcal{A}(S)}(\mathcal{A}(S)\in B).
	\end{align*}
	
	Since $\varepsilon > 0$, we have $e^{-\varepsilon}<1$. Therefore,
	\begin{align}
	\label{eq:rightside}
	(e^{-\varepsilon}-1)\le \mathbb P_{\mathcal{A}(S')}(\mathcal{A}(S')\in B)-\mathbb P_{\mathcal{A}(S)}(\mathcal{A}(S)\in B).
	\end{align}
	Eq. (\ref{eq:rightside}) stands for every neighbor sample set pair $S$ and $S'$. Thus,
	\begin{align*}
	& e^{-\varepsilon}-1 
	 \le \min\limits_{S\text{ and }S'\text{ neighbor}}(\mathbb P_{\mathcal{A}(S')}(\mathcal{A}(S')\in B)-\mathbb P_{\mathcal{A}(S)}(\mathcal{A}(S)\in B)).
	\end{align*}
	Therefore,
	\begin{align*}
	&\max\limits_{S\text{ and }S'\text{ neighbor}}\left\vert\left[\mathbb P_{\mathcal{A}(S')}(\mathcal{A}(S')\in B)-\mathbb P_{\mathcal{A}(S)}(\mathcal{A}(S)\in B)\right]\right\vert 
	\le 1-e^{-\varepsilon}.
	\end{align*}
	Thus, 
	\begin{align*}
	&\vert\mathbb{E}_{\mathcal{A}(S')}\ell(\mathcal{A}(S'),z)- \mathbb{E}_{\mathcal{A}(S)}\ell(\mathcal{A}(S),z)\vert
	\\
	=&\left\vert\int \ell(\mathcal{A}(S),z) \text{ }d\mathbb P_{\mathcal{A}(S)}-\int \ell(\mathcal{A}(S'),z) \text{d }\mathbb P_{\mathcal{A}(S')}\right\vert
	\\
	\overset{(*)}{\le}& \max\left\{ I_1, I_2 \right\}
	\\
	\le& M(1-e^{-\varepsilon}),
	\end{align*}
	where $I_1$ and $I_2$ in inequality $(*)$ is defined as 
	\begin{gather*}
	I_1 = \int_{\mathbb P_{\mathcal{A}(S)}>\mathbb P_{\mathcal{A}(S')}}\ell(\mathcal{A}(S),z) \left(\text{d}\mathbb P_{\mathcal{A}(S)}-\text{d}\mathbb P_{\mathcal{A}(S')} \right),\\
	I_2 = \int_{\mathbb P_{\mathcal{A}(S)}\le \mathbb P_{\mathcal{A}(S')}}\ell(\mathcal{A}(S),z)\left(\text{d}\mathbb P_{\mathcal{A}(S')}-\text{d}\mathbb P_{\mathcal{A}(S)}\right).
	\end{gather*}
	
	The proof is completed.
\end{proof}

Then, we prove Lemma \ref{lem:privacy_stability} using a different method.
\begin{proof}[Proof of Lemma \ref{lem:privacy_stability}]
	
	As the algorithm $\mathcal A$ is $(\varepsilon,\delta)$-differentially private, we have	
	\begin{equation*}
	\mathbb P_{\mathcal{A}(S)}(\mathcal{A}(S)\in B) \le e^\varepsilon \mathbb P_{\mathcal{A}(S')}(\mathcal{A}(S')\in B)+\delta,
	\end{equation*}
	where the subset $B$ is arbitrary from the hypothesis space $\mathcal H$. Let $B=\{h\in H: \ell(h,z)>t\}$. Then we have the following inequality,
	\begin{equation}
	\label{eq:dp>t}
	\mathbb P_{\mathcal{A}(S)}(\ell(\mathcal{A}(S),z)>t)\le e^{\varepsilon}\mathbb P_{\mathcal{A}(S')}(\ell(\mathcal{A}(S'),z)>t)+\delta.
	\end{equation}
	
	Additionally, $\mathbb{E}_{\mathcal{A}(S)} \ell(\mathcal{A}(S),z)$ is calculated as follows,
	\begin{align*}
	\mathbb{E}_{\mathcal{A}(S)} \ell(\mathcal{A}(S),z)=\int_{0}^{M} \mathbb P_{\mathcal{A}(S)}(\ell(\mathcal{A}(S),z)>t) dt.
	\end{align*}
	Applying eq. (\ref{eq:dp>t}), we have
	\begin{align}
	\label{eq:expectation_risk_proof_4}
	& \mathbb{E}_{\mathcal{A}(S)}\ell(\mathcal{A}(S),z) \nonumber \\
	=& \int_{0}^{M} \mathbb P_{\mathcal{A}(S)}(\ell(\mathcal{A}(S),z)>t) dt \nonumber\\
	\le& e^{\varepsilon}\int_{0}^{M} \mathbb P_{\mathcal{A}(S')}(\ell(\mathcal{A}(S'),z)>t) dt +M\delta \nonumber\\
	=& e^{\varepsilon}\mathbb{E}_{\mathcal{A}(S')} \ell (\mathcal{A}(S'),z) +M\delta.
	\end{align} 
	Rearranging eq.  (\ref{eq:expectation_risk_proof_4}), we have
	\begin{gather*}
	\nonumber
	e^{-\varepsilon} \mathbb{E}_{\mathcal{A}(S)}\ell (\mathcal{A}(S),z)
	\le  \mathbb{E}_{\mathcal{A}(S')}\ell (\mathcal{A}(S'),z)+e^{-\varepsilon} M\delta,
	\end{gather*}
	and
	\begin{align*}
	& (e^{-\varepsilon}-1) \mathbb{E}_{\mathcal{A}(S)}\ell (\mathcal{A}(S),z) 
	\le \mathbb{E}_{\mathcal{A}(S')}\ell (\mathcal{A}(S'),z)- \mathbb{E}_{\mathcal{A}(S)}\ell (\mathcal{A}(S),z)+e^{-\varepsilon} M\delta \label{eq:arbi}.
	\end{align*}
	Therefore,
	\begin{align*}
	&\mathbb{E}_{\mathcal{A}(S')}\ell (\mathcal{A}(S'),z)- \mathbb{E}_{\mathcal{A}(S)}\ell (\mathcal{A}(S),z) 
	\le e^{-\varepsilon} M\delta + (1-e^{-\varepsilon}) \mathbb{E}_{\mathcal{A}(S)}\ell (\mathcal{A}(S),z).
	\end{align*}
	
	Similarly, we can get the following inequality,
	\begin{align*}
	& - \mathbb{E}_{\mathcal{A}(S')}\ell (\mathcal{A}(S'),z) + \mathbb{E}_{\mathcal{A}(S)}\ell (\mathcal{A}(S),z) 
	\le e^{-\varepsilon} M\delta + (1-e^{-\varepsilon}) \mathbb{E}_{\mathcal{A}(S)}\ell (\mathcal{A}(S),z).
	\end{align*}
	
	Thus,
	\begin{align*}
	& \left\vert \mathbb E_{\mathcal{A}(S)}\ell (\mathcal{A}(S), Z) - \mathbb E_{\mathcal{A}(S')}\ell (\mathcal{A}(S'), Z) \right\vert 
	\le M \delta e^{-\varepsilon} + M(1-e^{-\varepsilon}).
	\end{align*}

	
	The proof is completed.
\end{proof}

\section{Experimental implementation details}
\label{sec:exp}

This section presents implementation details of our experiments, including the settings of adversarial training, the calculation of empirical composite robustified intensity and gradient noise, and the implementation of membership inference attack.
The running time of both ERM and adversarial training is also presented. The training code, learned models, and collected data are available at \url{https://github.com/fshp971/RPG}.



\subsection{Datasets}

We use the CIFAR-10 and CIFAR-100 datasets \cite{krizhevsky2009learning} in our experiments.
Both datasets contain $60,000$ $32 \times 32$ color images, where $50,000$ are training images and $10,000$ are test images. The images in CIFAR-10 are grouped into $10$ different categories, while images in CIFAR-100 have $100$ different categories. No pre-processing is involved. The datasets can be downloaded from \url{https://www.cs.toronto.edu/~kriz/cifar.html}.

\subsection{Neural network architectures}
\label{app:net}

We use a $34$-layer Wide ResNet \cite{zagoruyko2016wide}, WRN-34-10, in all experiments.
The detailed architectures of WRN-34-10 are presented in Table \ref{table:network}, where ``conv $x$-$c$'' represents a convolutional layer with kernel size $x \times x$ and $c$ output channels, ``fc-$c$'' represents a fully-connected layer with $c$ output channels, and ``$[\cdot]$'' represents the residual block named basic block \cite{he2016deep}.
Each convolution layer is followed by batch normalization and then ReLU activation.

\begin{table}[h]
\renewcommand\arraystretch{1.2}
\centering
\caption{Neural network architecture of WRN-34-10.}
 \begin{tabular}{c}
\toprule
WRN-34-10 \\
\midrule
conv3-16 \\
$\left[ \begin{matrix} \text{conv3-160} \\ \text{conv3-160} \end{matrix} \right]$ $\times$ 5 \\
$\left[ \begin{matrix} \text{conv3-320} \\ \text{conv3-320} \end{matrix} \right]$ $\times$ 5 \\
$\left[ \begin{matrix} \text{conv3-640} \\ \text{conv3-640} \end{matrix} \right]$ $\times$ 5 \\
avgpool \\
\midrule
fc-10 or fc-100 \\
\bottomrule
\end{tabular}
\label{table:network}
\end{table}

\subsection{Projected gradient descent}
\label{app:pgd}

Projected gradient descent (PGD) \cite{madry2018towards} performs $K$ iterative gradient ascent to search an example that maximizes the training loss. When using the $L_\infty$ norm, the $k$-th update in PGD is as follows,
\begin{equation*}
x_{k} = \prod_{\| x^\prime - x \|_\infty \le \rho} \left[ x_{k-1} + \alpha \cdot \text{sign} \left( \nabla_x l\left(h_\theta(x_{k-1}), y\right) \right) \right],
\end{equation*}
where $x_k$ is the adversarial example obtained in the $k$-th iteration, $\alpha$ is the step size, and $\prod_{x^\prime: \| x^\prime - x \|_\infty \le \rho}$ means that the projection is calculated in the ball sphere $ \mathbb B(x,\rho) = \{ x' : \| x' - x \|_\infty \le \rho \}$.
Besides, when using $L_2$ norm as the metric, the $k$-th update is as follows,
\begin{gather*}
x_{k} = \prod_{x^\prime: \| x^\prime - x \|_2 \le \rho} \left[ x_{k-1} + \alpha \cdot \nabla_x l\left(h_\theta(x_{k-1}), y\right) \right].
\end{gather*}

The iterations number $K$ is set as $8$. The step size $\alpha$ is set as $\rho/4$.

\subsection{Training settings}
\label{app:set}
All the models are trained by SGD for $80,000$ iterations. The momentum factor is set as $0.9$, the weight decay factor is set as $0.0002$, and the batch size is set as $128$. The learning rate is initialized as $0.1$ and then decays by $0.1$ every $30,000$ iterations. The list of the radius $\rho$ is presented in Table \ref{table:settings} as well as other factors.

\begin{table}[h]
    \renewcommand\arraystretch{1.5}
    \centering
    \caption{Details of different experimental settings.}
\begin{tabular}{c c c c}
\toprule
\multirow{2}{1cm}{\centering Setting} & \multirow{2}{1cm}{\centering Dataset} & \multirow{2}{1cm}{\centering Norm of PGD} & \multirow{2}{1.5cm}{\centering Radius $\rho$ } \\ \\
\hline
A & CIFAR-10 & $L_\infty$ & $\{\frac{i}{255} : i = 0, \ 1, \ \dots, \ 10\}$ \\
B & CIFAR-100 & $L_\infty$ & $\{\frac{i}{255} : i = 0, \ 1, \ \dots, \ 10\}$ \\
C & CIFAR-10 & $L_2$ & $\{\frac{25 \cdot i}{255} : i = 0, \ 1, \ \cdots, \ 12 \}$ \\
D & CIFAR-100 & $L_2$ & $\{\frac{25 \cdot i}{255} : i = 0, \ 1, \ \cdots, \ 12 \}$ \\
\bottomrule
\end{tabular}
\label{table:settings}
\end{table}

\subsection{Calculation of empirical robustified intensity}

Calculating the empirical robustified intensity would be of a considerable high computational cost. We thus apply a sparse version in our experiments. We calculate the empirical single-iteration robustified intensities every $m$ iterations. Thus, we obtain $\lfloor T/m \rfloor$ empirical single-iteration robustified intensities $\hat I_{m}, \hat I_{2m}, \cdots, \hat I_{\lfloor T/m \rfloor \cdot m}$. The sparse version of empirical robustified intensity is as below,
\begin{gather*}
	\left( \frac{1}{\lfloor T/m \rfloor} \sum_{i=1}^{\lfloor T/m \rfloor} \hat I_{m \cdot i}^4 \right)^{\frac{1}{4}}.
\end{gather*}
In our experiments, the calculation interval $m$ is set as $100$.


\subsection{Calculation of gradient noise}

We apply a five-step approach to estimate the gradient noise for a specific parameter $\theta$: (1) calculate the gradient $\nabla_\theta \mathcal{\hat R}_S(\theta)$ of the full training set $S$; (2) randomly draw a mini batch from the training set and then calculate the gradient whereon; (3) calculate the difference  between the mini-batch gradient and its full-batch counterpart; (4) randomly sample a  set of components of the difference vector in the previous step; and (5) normalize all components in the sampled vector by their standard deviation.

\subsection{Membership inference attack}
\label{app:attack}

We employ a  threshold-based version of membership inference attack \cite{yeom2018privacy} to empirically assess the privacy-preserving abilities.
Given a training set $S_{\text{train}}$, a test set $S_{\text{test}}$, and a trained model $h_\theta(\cdot)$. Suppose a data point $(x,y)$ comes from $S_{\text{train}}$ or $S_{\text{test}}$ with equal probabilities. Then, the membership inference attack accuracy with a threshold $\zeta$ is calculated as follows,
\begin{align*}
Acc(\zeta) &= \frac{1}{2} \times
\left( \frac{\sum_{(x,y) \in S_{\text{train}}} \bm{1}[h_{\theta}(x)_y \geq \zeta]}{\vert S_{\text{train}} \vert} 
+ \frac{\sum_{(x,y) \in S_{\text{test}}} \bm{1}[h_{\theta}(x)_y < \zeta]}{\vert S_{\text{test}} \vert} \right),
\end{align*}
where $h_\theta(x)_y$ is the output confidence for label $y$ and $\bm{1}[\cdot]$ is the indicator function. Therefore, the goal of the threshold-based attack model is to find an optimal threshold $\zeta_{\text{optim}}$ that maximizes the attack accuracy, {\it i.e.},
\begin{equation*}
\zeta_{\text{optim}} = \arg \max_{\zeta} Acc(\zeta),
\end{equation*}
and this can be done by enumerating all possible threshold values $\zeta$.

\subsection{Hardware environment}

All our experiments are conducted on a computing cluster with GPUs of NVIDIA\textsuperscript{\textregistered} Tesla\textsuperscript{\texttrademark} V100 16GB and CPUs of Intel\textsuperscript{\textregistered} Xeon\textsuperscript{\textregistered} Gold 6140 CPU @ 2.30GHz.

\subsection{Running time}
\label{app:runtime}

We estimate the experiment running time based on the log files, as shown in Table \ref{table:runtime}.
The running time includes both the training time and the time of calculating the robustified intensities in every setting.
It is worth noting that precise running time may vary depending on the specific experimental conditions such as the temperature of GPUs and the working load of computing cluster. Therefore, the estimated running time given here would be not precise.

\begin{table}[h]
    \renewcommand\arraystretch{1.5}
    \centering
    \caption{Estimated running time of each experimental setting. ``AT'' is for adversarial training.}
    \begin{tabular}{c c c c c}
        \toprule
        & Setting A & Setting B & Setting C & Setting D \\
        \hline
        ERM & 7.5 hrs & 7.4 hrs & 7.5 hrs & 7.4 hrs \\
        AT & 53.1 hrs & 49.8 hrs & 56.4 hrs & 52.4 hrs \\
        \bottomrule
    \end{tabular}
\label{table:runtime}
\end{table}

\section{Conclusion}
\label{sec:conclusion}

This paper studies the privacy-preserving and generalization abilities of adversarial training. We prove that the adversarial robustness, privacy preservation, and generalization are interrelated from both theoretical and empirical perspectives. We define {\it robustified intensity} and design its empirical version, {\it empirical robustified intensity}, which is proved to be asymptotically consistent with the robustified intensity. We then prove that adversarial training scheme is $(\varepsilon, \delta)$-differentially private, in which the magnitude of the differential privacy $(\varepsilon, \delta)$ has a positive correlation with the robustified intensity. Based on the privacy-robustness relationship, an $\mathcal O(\sqrt{\log N}/N)$ on-average generalization bound and a $\mathcal O(1/\sqrt{N})$ high-probability one for adversarial training are delivered ($N$ is the training sample size). Extensive systematic experiments are conducted based on network architecture Wide ResNet and datasets CIFAR-10, CIFAR-100. The results fully support our theories.

\bibliographystyle{plain}
\bibliography{bib}

\begin{thebibliography}{10}

\bibitem{abadi2016deep}
Martin Abadi, Andy Chu, Ian Goodfellow, H~Brendan McMahan, Ilya Mironov, Kunal
  Talwar, and Li~Zhang.
\newblock Deep learning with differential privacy.
\newblock In {\em ACM SIGSAC Conference on Computer and Communications
  Security}, pages 308--318, 2016.

\bibitem{arachchige2019local}
Pathum Chamikara~Mahawaga Arachchige, Peter Bertok, Ibrahim Khalil, Dongxi Liu,
  Seyit Camtepe, and Mohammed Atiquzzaman.
\newblock Local differential privacy for deep learning.
\newblock {\em IEEE Internet of Things Journal}, 2019.

\bibitem{baluja2018learning}
Shumeet Baluja and Ian Fischer.
\newblock Learning to attack: Adversarial transformation networks.
\newblock In {\em AAAI Conference on Artificial Intelligence}, volume~1,
  page~3, 2018.

\bibitem{bartlett2017spectrally}
Peter~L Bartlett, Dylan~J Foster, and Matus~J Telgarsky.
\newblock Spectrally-normalized margin bounds for neural networks.
\newblock In {\em Advances in Neural Information Processing Systems}, pages
  6240--6249, 2017.

\bibitem{bartlett2002rademacher}
Peter~L Bartlett and Shahar Mendelson.
\newblock Rademacher and {Gaussian} complexities: Risk bounds and structural
  results.
\newblock {\em Journal of Machine Learning Research}, 3(Nov):463--482, 2002.

\bibitem{biggio2013evasion}
Battista Biggio, Igino Corona, Davide Maiorca, Blaine Nelson, Nedim
  {\v{S}}rndi{\'c}, Pavel Laskov, Giorgio Giacinto, and Fabio Roli.
\newblock Evasion attacks against machine learning at test time.
\newblock In {\em European Conference on Machine Learning}, 2013.

\bibitem{blumer1989learnability}
Anselm Blumer, Andrzej Ehrenfeucht, David Haussler, and Manfred~K Warmuth.
\newblock Learnability and the {Vapnik-Chervonenkis} dimension.
\newblock {\em Journal of the ACM}, 36(4):929--965, 1989.

\bibitem{bousquet2002stability}
Olivier Bousquet and Andr{\'e} Elisseeff.
\newblock Stability and generalization.
\newblock {\em Journal of Machine Learning Research}, 2(Mar):499--526, 2002.

\bibitem{bun2016concentrated}
Mark Bun and Thomas Steinke.
\newblock Concentrated differential privacy: Simplifications, extensions, and
  lower bounds.
\newblock In {\em Theory of Cryptography Conference}, pages 635--658, 2016.

\bibitem{chaudhuri2019capacity}
Kamalika Chaudhuri, Jacob Imola, and Ashwin Machanavajjhala.
\newblock Capacity bounded differential privacy.
\newblock {\em arXiv preprint arXiv:1907.02159}, 2019.

\bibitem{chen2018rise}
Hongming Chen, Ola Engkvist, Yinhai Wang, Marcus Olivecrona, and Thomas
  Blaschke.
\newblock The rise of deep learning in drug discovery.
\newblock {\em Drug Discovery Today}, 23(6):1241--1250, 2018.

\bibitem{chen2020universal}
Sizhe Chen, Zhengbao He, Chengjin Sun, Jie Yang, and Xiaolin Huang.
\newblock Universal adversarial attack on attention and the resulting dataset
  damagenet.
\newblock {\em IEEE Transactions on Pattern Analysis and Machine Intelligence},
  2020.

\bibitem{cuff2016differential}
Paul Cuff and Lanqing Yu.
\newblock Differential privacy as a mutual information constraint.
\newblock In {\em ACM SIGSAC Conference on Computer and Communications
  Security}, pages 43--54, 2016.

\bibitem{dai2018adversarial}
Hanjun Dai, Hui Li, Tian Tian, Xin Huang, Lin Wang, Jun Zhu, and Le~Song.
\newblock Adversarial attack on graph structured data.
\newblock {\em arXiv preprint arXiv:1806.02371}, 2018.

\bibitem{dudley1967sizes}
Richard~M Dudley.
\newblock The sizes of compact subsets of hilbert space and continuity of
  {Gaussian} processes.
\newblock {\em Journal of Functional Analysis}, 1(3):290--330, 1967.

\bibitem{dwork2015generalization}
Cynthia Dwork, Vitaly Feldman, Moritz Hardt, Toni Pitassi, Omer Reingold, and
  Aaron Roth.
\newblock Generalization in adaptive data analysis and holdout reuse.
\newblock In {\em Advances in Neural Information Processing Systems}, pages
  2350--2358, 2015.

\bibitem{dwork2013s}
Cynthia Dwork and Deirdre~K Mulligan.
\newblock It's not privacy, and it's not fair.
\newblock {\em Stanford Law Review Online}, 66:35, 2013.

\bibitem{dwork2014algorithmic}
Cynthia Dwork and Aaron Roth.
\newblock The algorithmic foundations of differential privacy.
\newblock {\em Foundations and Trends{\textregistered} in Theoretical Computer
  Science}, 9(3--4):211--407, 2014.

\bibitem{dwork2016concentrated}
Cynthia Dwork and Guy~N Rothblum.
\newblock Concentrated differential privacy.
\newblock {\em arXiv preprint arXiv:1603.01887}, 2016.

\bibitem{e2020towards}
Weinan E, Chao Ma, Stephan Wojtowytsch, and Lei Wu.
\newblock Towards a mathematical understanding of neural network-based machine
  learning: What we know and what we don't.
\newblock {\em arXiv preprint arXiv:2009.10713}, 2020.

\bibitem{feldman2019high}
Vitaly Feldman and Jan Vondrak.
\newblock High probability generalization bounds for uniformly stable
  algorithms with nearly optimal rate.
\newblock {\em arXiv preprint arXiv:1902.10710}, 2019.

\bibitem{fischer2018deep}
Thomas Fischer and Christopher Krauss.
\newblock Deep learning with long short-term memory networks for financial
  market predictions.
\newblock {\em European Journal of Operational Research}, 270(2):654--669,
  2018.

\bibitem{geumlek2017renyi}
Joseph Geumlek, Shuang Song, and Kamalika Chaudhuri.
\newblock Renyi differential privacy mechanisms for posterior sampling.
\newblock In {\em Advances in Neural Information Processing Systems}, pages
  5289--5298, 2017.

\bibitem{gilmer2018motivating}
Justin Gilmer, Ryan~P Adams, Ian Goodfellow, David Andersen, and George~E Dahl.
\newblock Motivating the rules of the game for adversarial example research.
\newblock {\em arXiv preprint arXiv:1807.06732}, 2018.

\bibitem{golowich2017size}
Noah Golowich, Alexander Rakhlin, and Ohad Shamir.
\newblock Size-independent sample complexity of neural networks.
\newblock In {\em Annual Conference on Learning Theory}, pages 297--299, 2018.

\bibitem{goodfellow2014explaining}
Ian~J Goodfellow, Jonathon Shlens, and Christian Szegedy.
\newblock Explaining and harnessing adversarial examples.
\newblock {\em arXiv preprint arXiv:1412.6572}, 2014.

\bibitem{hardt2016train}
Moritz Hardt, Ben Recht, and Yoram Singer.
\newblock Train faster, generalize better: Stability of stochastic gradient
  descent.
\newblock In {\em International Conference on Machine learning}, pages
  1225--1234, 2016.

\bibitem{harvey2017nearly}
Nick Harvey, Christopher Liaw, and Abbas Mehrabian.
\newblock Nearly-tight {VC}-dimension bounds for piecewise linear neural
  networks.
\newblock In {\em Annual Conference on Learning Theory}, pages 1064--1068,
  2017.

\bibitem{haussler1995sphere}
David Haussler.
\newblock Sphere packing numbers for subsets of the boolean $n$-cube with
  bounded {Vapnik-Chervonenkis} dimension.
\newblock {\em Journal of Combinatorial Theory, Series A}, 69(2):217--232,
  1995.

\bibitem{he2019control}
Fengxiang He, Tongliang Liu, and Dacheng Tao.
\newblock Control batch size and learning rate to generalize well: Theoretical
  and empirical evidence.
\newblock In {\em Advances in Neural Information Processing Systems}, 2019.

\bibitem{he2020recent}
Fengxiang He and Dacheng Tao.
\newblock Recent advances in deep learning theory.
\newblock {\em arXiv preprint arXiv:2012.10931}, 2020.

\bibitem{he2020tighter}
Fengxiang He, Bohan Wang, and Dacheng Tao.
\newblock Tighter generalization bounds for iterative differentially private
  learning algorithms.
\newblock {\em arXiv preprint arXiv:2007.09371}, 2020.

\bibitem{he2016deep}
Kaiming He, Xiangyu Zhang, Shaoqing Ren, and Jian Sun.
\newblock Deep residual learning for image recognition.
\newblock In {\em IEEE Conference on Computer Vision and Pattern Recognition},
  2016.

\bibitem{jastrzkebski2017three}
Stanis{\l}aw Jastrzebski, Zachary Kenton, Devansh Arpit, Nicolas Ballas, Asja
  Fischer, Yoshua Bengio, and Amos Storkey.
\newblock Three factors influencing minima in sgd.
\newblock {\em arXiv preprint arXiv:1711.04623}, 2017.

\bibitem{karim2020adversarial}
Fazle Karim, Somshubra Majumdar, and Houshang Darabi.
\newblock Adversarial attacks on time series.
\newblock {\em IEEE Transactions on Pattern Analysis and Machine Intelligence},
  2020.

\bibitem{kearns1999algorithmic}
Michael Kearns and Dana Ron.
\newblock Algorithmic stability and sanity-check bounds for leave-one-out
  cross-validation.
\newblock {\em Neural Computation}, 11(6):1427--1453, 1999.

\bibitem{khim2018adversarial}
Justin Khim and Po-Ling Loh.
\newblock Adversarial risk bounds via function transformation.
\newblock {\em arXiv preprint arXiv:1810.09519}, 2018.

\bibitem{kingma2014adam}
Diederik~P Kingma and Jimmy Ba.
\newblock Adam: A method for stochastic optimization.
\newblock {\em arXiv preprint arXiv:1412.6980}, 2014.

\bibitem{koltchinskii2001rademacher}
Vladimir Koltchinskii.
\newblock Rademacher penalties and structural risk minimization.
\newblock {\em IEEE Transactions on Information Theory}, 47(5):1902--1914,
  2001.

\bibitem{koltchinskii2000rademacher}
Vladimir Koltchinskii and Dmitriy Panchenko.
\newblock Rademacher processes and bounding the risk of function learning.
\newblock In {\em High Dimensional Probability II}, pages 443--457. Springer,
  2000.

\bibitem{krizhevsky2009learning}
Alex Krizhevsky and Geoffrey Hinton.
\newblock Learning multiple layers of features from tiny images.
\newblock 2009.

\bibitem{kulikowski1980artificial}
Casimir~A Kulikowski.
\newblock Artificial intelligence methods and systems for medical consultation.
\newblock {\em IEEE Transactions on Pattern Analysis and Machine Intelligence},
  (5):464--476, 1980.

\bibitem{kurakin2016adversarial}
Alexey Kurakin, Ian Goodfellow, and Samy Bengio.
\newblock Adversarial examples in the physical world.
\newblock {\em arXiv preprint arXiv:1607.02533}, 2016.

\bibitem{kushner2003stochastic}
Harold Kushner and G~George Yin.
\newblock {\em Stochastic approximation and recursive algorithms and
  applications}, volume~35.
\newblock Springer Science \& Business Media, 2003.

\bibitem{kuzborskij2018data}
Ilja Kuzborskij and Christoph Lampert.
\newblock Data-dependent stability of stochastic gradient descent.
\newblock In {\em International Conference on Machine Learning}, pages
  2815--2824, 2018.

\bibitem{lecuyer2018connection}
Mathias Lecuyer, Vaggelis Atlidakis, Roxana Geambasu, Daniel Hsu, and Suman
  Jana.
\newblock On the connection between differential privacy and adversarial
  robustness in machine learning.
\newblock {\em arXiv preprint arXiv:1802.03471}, 2018.

\bibitem{lecuyer2019certified}
Mathias Lecuyer, Vaggelis Atlidakis, Roxana Geambasu, Daniel Hsu, and Suman
  Jana.
\newblock Certified robustness to adversarial examples with differential
  privacy.
\newblock In {\em IEEE Symposium on Security and Privacy}, pages 656--672,
  2019.

\bibitem{li2018second}
Bai Li, Changyou Chen, Wenlin Wang, and Lawrence Carin.
\newblock Second-order adversarial attack and certifiable robustness.
\newblock 2018.

\bibitem{liang2019fisher}
Tengyuan Liang, Tomaso Poggio, Alexander Rakhlin, and James Stokes.
\newblock Fisher-rao metric, geometry, and complexity of neural networks.
\newblock In {\em International Conference on Artificial Intelligence and
  Statistics}, pages 888--896, 2019.

\bibitem{liao2017hypothesis}
Jiachun Liao, Lalitha Sankar, Vincent~YF Tan, and Flavio du~Pin~Calmon.
\newblock Hypothesis testing under mutual information privacy constraints in
  the high privacy regime.
\newblock {\em IEEE Transactions on Information Forensics and Security},
  13(4):1058--1071, 2017.

\bibitem{litjens2017survey}
Geert Litjens, Thijs Kooi, Babak~Ehteshami Bejnordi, Arnaud Arindra~Adiyoso
  Setio, Francesco Ciompi, Mohsen Ghafoorian, Jeroen~Awm Van Der~Laak, Bram
  Van~Ginneken, and Clara~I S{\'a}nchez.
\newblock A survey on deep learning in medical image analysis.
\newblock {\em Medical Image Analysis}, 42:60--88, 2017.

\bibitem{ljung2012stochastic}
Lennart Ljung, Georg Pflug, and Harro Walk.
\newblock {\em Stochastic approximation and optimization of random systems},
  volume~17.
\newblock Birkh{\"a}user, 2012.

\bibitem{madry2018towards}
Aleksander Madry, Aleksandar Makelov, Ludwig Schmidt, Dimitris Tsipras, and
  Adrian Vladu.
\newblock Towards deep learning models resistant to adversarial attacks.
\newblock In {\em International Conference on Learning Representations}, 2018.

\bibitem{mandt2017stochastic}
Stephan Mandt, Matthew~D Hoffman, and David~M Blei.
\newblock Stochastic gradient descent as approximate {Bayesian} inference.
\newblock {\em Journal of Machine Learning Research}, 18(1):4873--4907, 2017.

\bibitem{mcallester1999pac}
David~A McAllester.
\newblock {PAC-Bayesian} model averaging.
\newblock In {\em Annual Conference of Learning Theory}, volume~99, pages
  164--170, 1999.

\bibitem{mcallester1999some}
David~A McAllester.
\newblock Some {PAC-Bayesian} theorems.
\newblock {\em Machine Learning}, 37(3):355--363, 1999.

\bibitem{mironov2017renyi}
Ilya Mironov.
\newblock R{\'e}nyi differential privacy.
\newblock In {\em IEEE Computer Security Foundations Symposium}, pages
  263--275, 2017.

\bibitem{mohri2018foundations}
Mehryar Mohri, Afshin Rostamizadeh, and Ameet Talwalkar.
\newblock {\em Foundations of machine learning}.
\newblock MIT press, 2018.

\bibitem{mou2018generalization}
Wenlong Mou, Liwei Wang, Xiyu Zhai, and Kai Zheng.
\newblock Generalization bounds of sgld for non-convex learning: Two
  theoretical viewpoints.
\newblock In {\em Annual Conference On Learning Theory}, pages 605--638, 2018.

\bibitem{nakkiran2019adversarial}
Preetum Nakkiran.
\newblock Adversarial robustness may be at odds with simplicity.
\newblock {\em arXiv preprint arXiv:1901.00532}, 2019.

\bibitem{nesterov1983method}
Yurii~E Nesterov.
\newblock A method for solving the convex programming problem with convergence
  rate o (1/k\^{} 2).
\newblock In {\em Dokl. Akad. Nauk Sssr}, volume 269, pages 543--547, 1983.

\bibitem{neyshabur2017pac}
Behnam Neyshabur, Srinadh Bhojanapalli, and Nathan Srebro.
\newblock A {PAC-Bayesian} approach to spectrally-normalized margin bounds for
  neural networks.
\newblock {\em arXiv preprint arXiv:1707.09564}, 2017.

\bibitem{nguyen2015deep}
Anh Nguyen, Jason Yosinski, and Jeff Clune.
\newblock Deep neural networks are easily fooled: High confidence predictions
  for unrecognizable images.
\newblock In {\em IEEE Conference on Computer Vision and Pattern Recognition},
  pages 427--436, 2015.

\bibitem{papernot2017practical}
Nicolas Papernot, Patrick McDaniel, Ian Goodfellow, Somesh Jha, Z~Berkay Celik,
  and Ananthram Swami.
\newblock Practical black-box attacks against machine learning.
\newblock In {\em ACM on Asia Conference on Computer and Communications
  Security}, pages 506--519, 2017.

\bibitem{papernot2016limitations}
Nicolas Papernot, Patrick McDaniel, Somesh Jha, Matt Fredrikson, Z~Berkay
  Celik, and Ananthram Swami.
\newblock The limitations of deep learning in adversarial settings.
\newblock In {\em IEEE European Symposium on Security and Privacy}, pages
  372--387, 2016.

\bibitem{phan2019preserving}
NhatHai Phan, Ruoming Jin, My~T Thai, Han Hu, and Dejing Dou.
\newblock Preserving differential privacy in adversarial learning with provable
  robustness.
\newblock {\em arXiv preprint arXiv:1903.09822}, 2019.

\bibitem{pinot2019unified}
Rafael Pinot, Florian Yger, C{\'e}dric Gouy-Pailler, and Jamal Atif.
\newblock A unified view on differential privacy and robustness to adversarial
  examples.
\newblock {\em arXiv preprint arXiv:1906.07982}, 2019.

\bibitem{poggio2020theoretical}
Tomaso Poggio, Andrzej Banburski, and Qianli Liao.
\newblock Theoretical issues in deep networks.
\newblock {\em Proceedings of the National Academy of Sciences}, 2020.

\bibitem{robbins1951stochastic}
Herbert Robbins and Sutton Monro.
\newblock A stochastic approximation method.
\newblock {\em The Annals of Mathematical Statistics}, pages 400--407, 1951.

\bibitem{rogers1978finite}
William~H Rogers and Terry~J Wagner.
\newblock A finite sample distribution-free performance bound for local
  discrimination rules.
\newblock {\em The Annals of Statistics}, pages 506--514, 1978.

\bibitem{schmidt2018adversarially}
Ludwig Schmidt, Shibani Santurkar, Dimitris Tsipras, Kunal Talwar, and
  Aleksander Madry.
\newblock Adversarially robust generalization requires more data.
\newblock {\em Advances in Neural Information Processing Systems},
  31:5014--5026, 2018.

\bibitem{shalev2010learnability}
Shai Shalev-Shwartz, Ohad Shamir, Nathan Srebro, and Karthik Sridharan.
\newblock Learnability, stability and uniform convergence.
\newblock {\em Journal of Machine Learning Research}, 11(Oct):2635--2670, 2010.

\bibitem{shokri2017membership}
Reza Shokri, Marco Stronati, Congzheng Song, and Vitaly Shmatikov.
\newblock Membership inference attacks against machine learning models.
\newblock In {\em IEEE Symposium on Security and Privacy (SP)}, pages 3--18,
  2017.

\bibitem{silver2016mastering}
David Silver, Aja Huang, Chris~J Maddison, Arthur Guez, Laurent Sifre, George
  Van Den~Driessche, Julian Schrittwieser, Ioannis Antonoglou, Veda
  Panneershelvam, and Marc Lanctot.
\newblock Mastering the game of go with deep neural networks and tree search.
\newblock {\em Nature}, 529(7587):484, 2016.

\bibitem{smith2018bayesian}
Samuel~L Smith and Quoc~V Le.
\newblock A {Bayesian} perspective on generalization and stochastic gradient
  descent.
\newblock In {\em International Conference on Learning Representations}, 2018.

\bibitem{snelick2005large}
Robert Snelick, Umut Uludag, Alan Mink, Mike Indovina, and Anil Jain.
\newblock Large-scale evaluation of multimodal biometric authentication using
  state-of-the-art systems.
\newblock {\em IEEE Transactions on Pattern Analysis and Machine Intelligence},
  27(3):450--455, 2005.

\bibitem{song2019privacy}
Liwei Song, Reza Shokri, and Prateek Mittal.
\newblock Privacy risks of securing machine learning models against adversarial
  examples.
\newblock In {\em ACM SIGSAC Conference on Computer and Communications
  Security}, pages 241--257, 2019.

\bibitem{sun2019towards}
Ke~Sun, Zhanxing Zhu, and Zhouchen Lin.
\newblock Towards understanding adversarial examples systematically: Exploring
  data size, task and model factors.
\newblock {\em arXiv preprint arXiv:1902.11019}, 2019.

\bibitem{sun2006road}
Zehang Sun, George Bebis, and Ronald Miller.
\newblock On-road vehicle detection: A review.
\newblock {\em IEEE Transactions on Pattern Analysis and Machine Intelligence},
  28(5):694--711, 2006.

\bibitem{szegedy2013intriguing}
Christian Szegedy, Wojciech Zaremba, Ilya Sutskever, Joan Bruna, Dumitru Erhan,
  Ian Goodfellow, and Rob Fergus.
\newblock Intriguing properties of neural networks.
\newblock {\em arXiv preprint arXiv:1312.6199}, 2013.

\bibitem{tseng1998incremental}
Paul Tseng.
\newblock An incremental gradient (-projection) method with momentum term and
  adaptive stepsize rule.
\newblock {\em SIAM Journal on Optimization}, 8(2):506--531, 1998.

\bibitem{tsipras2019robustness}
Dimitris Tsipras, Shibani Santurkar, Logan Engstrom, Alexander Turner, and
  Aleksander Madry.
\newblock Robustness may be at odds with accuracy.
\newblock In {\em International Conference on Learning Representations}, 2019.

\bibitem{tu2020understanding}
Zhuozhuo Tu, Fengxiang He, and Dacheng Tao.
\newblock Understanding generalization in recurrent neural networks.
\newblock In {\em International Conference on Learning Representations}, 2020.

\bibitem{tu2019theoretical}
Zhuozhuo Tu, Jingwei Zhang, and Dacheng Tao.
\newblock Theoretical analysis of adversarial learning: A minimax approach.
\newblock In {\em Advances in Neural Information Processing Systems}, pages
  12280--12290, 2019.

\bibitem{vapnik2006estimation}
Vladimir Vapnik.
\newblock {\em Estimation of Dependences based on Empirical Data}.
\newblock Springer Science \& Business Media, 2006.

\bibitem{vapnik2013nature}
Vladimir Vapnik.
\newblock {\em The Nature of Statistical Learning Theory}.
\newblock Springer Science \& Business Media, 2013.

\bibitem{verma2019stability}
Saurabh Verma and Zhi-Li Zhang.
\newblock Stability and generalization of graph convolutional neural networks.
\newblock In {\em ACM SIGKDD International Conference on Knowledge Discovery \&
  Data Mining}, pages 1539--1548, 2019.

\bibitem{wang2012framework}
Fei Wang, Noah Lee, Jianying Hu, Jimeng Sun, Shahram Ebadollahi, and Andrew~F
  Laine.
\newblock A framework for mining signatures from event sequences and its
  applications in healthcare data.
\newblock {\em IEEE Transactions on Pattern Analysis and Machine Intelligence},
  35(2):272--285, 2012.

\bibitem{wang2020hamiltonian}
Hongjun Wang, Guanbin Li, Xiaobai Liu, and Liang Lin.
\newblock A hamiltonian monte carlo method for probabilistic adversarial attack
  and learning.
\newblock {\em IEEE Transactions on Pattern Analysis and Machine Intelligence},
  2020.

\bibitem{wang2016relation}
Weina Wang, Lei Ying, and Junshan Zhang.
\newblock On the relation between identifiability, differential privacy, and
  mutual-information privacy.
\newblock {\em IEEE Transactions on Information Theory}, 62(9):5018--5029,
  2016.

\bibitem{wen2019geometry}
Yuxin Wen, Jiehong Lin, Ke~Chen, CL~Philip Chen, and Kui Jia.
\newblock Geometry-aware generation of adversarial point clouds.
\newblock {\em IEEE Transactions on Pattern Analysis and Machine Intelligence},
  2019.

\bibitem{xu2011sparse}
Huan Xu, Constantine Caramanis, and Shie Mannor.
\newblock Sparse algorithms are not stable: A no-free-lunch theorem.
\newblock {\em IEEE Transactions on Pattern Analysis and Machine Intelligence},
  34(1):187--193, 2011.

\bibitem{yeom2018privacy}
Samuel Yeom, Irene Giacomelli, Matt Fredrikson, and Somesh Jha.
\newblock Privacy risk in machine learning: Analyzing the connection to
  overfitting.
\newblock In {\em IEEE Computer Security Foundations Symposium}, pages
  268--282, 2018.

\bibitem{yin2019rademacher}
Dong Yin, Ramchandran Kannan, and Peter Bartlett.
\newblock Rademacher complexity for adversarially robust generalization.
\newblock In {\em International Conference on Machine learning}, pages
  7085--7094, 2019.

\bibitem{zagoruyko2016wide}
Sergey Zagoruyko and Nikos Komodakis.
\newblock Wide residual networks.
\newblock {\em arXiv preprint arXiv:1605.07146}, 2016.

\bibitem{zheng2019distributionally}
Tianhang Zheng, Changyou Chen, and Kui Ren.
\newblock Distributionally adversarial attack.
\newblock In {\em AAAI Conference on Artificial Intelligence}, volume~33, pages
  2253--2260, 2019.

\end{thebibliography}

\end{document}